%% file: main.tex
\documentclass[10pt,twocolumn,letterpaper]{article}

\usepackage[utf8]{inputenc} 
\usepackage[T1]{fontenc}    
\usepackage{url}            
\usepackage{booktabs}       
\usepackage{amsfonts}       
\usepackage{nicefrac}       
\usepackage{microtype}      
\usepackage{xcolor}         
\usepackage{amssymb}
\usepackage{float}
\usepackage{amsmath}
\usepackage{amsthm}
\usepackage{graphicx}
\usepackage{diagbox}
\usepackage{tabularx}
\usepackage{bm}
\usepackage{bbm}
\usepackage{algorithm}
\usepackage{algorithmic}
\usepackage{setspace}
\usepackage{subfigure}

\usepackage{amsfonts}
\usepackage{float}
\usepackage{enumitem,graphicx}

\usepackage{xcolor}

\usepackage{multirow}

\def\beps{\bm{\epsilon}}

\def\w{\bm{w}}
\def\x{\bm{x}}
\def\y{\bm{y}}

\def\T{\mathbf{T}}
\def\beps{\boldsymbol{\epsilon}}

\def\Db{\mathcal{D}_{benign}}
\def\gL{\mathcal{L}}

\makeatletter
\DeclareRobustCommand\onedot{\futurelet\@let@token\@onedot}
\def\@onedot{\ifx\@let@token.\else.\null\fi\xspace}

\def\eg{\emph{e.g}\onedot} 

\def\ie{\emph{i.e}\onedot}

\def\wrt{\emph{w.r.t}\onedot} 

\def\etal{\emph{et al}\onedot}

\newcommand*{\email}[1]{%
    \normalsize\href{mailto:#1}{#1}\par
    }
\usepackage[numbers,square,comma]{natbib}

\usepackage{iccv}
\usepackage{times}
\usepackage{epsfig}
\usepackage{graphicx}
\usepackage{amsmath}
\usepackage{amssymb}
\usepackage{xcolor}

\usepackage[pagebackref=true,breaklinks=true,letterpaper=true,colorlinks,bookmarks=false]{hyperref}

\iccvfinalcopy 

\ificcvfinal\fi

\begin{document}

\title{Enhancing Fine-Tuning Based Backdoor Defense with Sharpness-Aware Minimization}

\author{
Mingli Zhu\textsuperscript{1}
\ \ \ \ Shaokui Wei\textsuperscript{1} 
\ \ \ \ Li Shen\textsuperscript{2} 
\ \ \ \ Yanbo Fan\textsuperscript{3}
\ \ \ \ Baoyuan Wu\textsuperscript{1}\thanks{Corresponds to Baoyuan Wu (\email{wubaoyuan@cuhk.edu.cn}).}
\\
\textsuperscript{1}School of Data Science, Shenzhen Research Institute of Big Data, \\
The Chinese University of Hong Kong, Shenzhen\\
\textsuperscript{2}JD Explore Academy \\
\textsuperscript{3}Tencent AI Lab
}


\maketitle


\long\def\comment#1{}
\newcommand{\ls}[1]{{\color{blue}{\bf\sf [LS: #1]}}}
\newcommand{\fan}[1]{{\color{red}{\bf\sf [Fan: #1]}}}
\newcommand{\ml}[1]{{\color{red}{\bf\sf [ML: #1]}}}
\newcommand{\wsk}[1]{{\color{red}{\bf\sf [Wei: #1]}}}

\begin{abstract}
Backdoor defense, which aims to detect or mitigate the effect of malicious triggers introduced by attackers, is becoming increasingly critical for machine learning security and integrity. Fine-tuning based on benign data is a natural defense to erase the backdoor effect in a backdoored model. However, recent studies show that, given limited benign data, vanilla fine-tuning has poor defense performance. 
In this work, we provide a deep study of fine-tuning the backdoored model from the neuron perspective and find that backdoor-related neurons fail to escape the local minimum in the fine-tuning process. 
Inspired by observing that the backdoor-related neurons often have larger norms, we propose FT-SAM, a novel backdoor defense paradigm that aims to shrink the norms of backdoor-related neurons by incorporating sharpness-aware minimization with fine-tuning. We demonstrate the effectiveness of our method on several benchmark datasets and network architectures, where it achieves state-of-the-art defense performance. Overall, our work provides a promising avenue for improving the robustness of machine learning models against backdoor attacks.

\end{abstract}

\input{contents/1_introduction}

\input{contents/2_relatedwork}
\input{contents/3_method}

\input{contents/4_experiment}

\input{contents/5_conclusion}

\clearpage
{\small
\bibliographystyle{ieee_fullname}
\bibliography{main}
}
\clearpage
\input{contents/appendix}

\end{document}

%% file: contents/1_introduction.tex
\section{Introduction\label{sec1}}

As deep neural networks (DNNs) have been increasingly applied to safety-critical tasks such as face recognition, autonomous driving, and medical image processing \cite{he2016deep, adjabi2020past, liu2020computing, tournier2019mrtrix3,xiao2018iot}, the threat exhibited by DNNs has drawn attention from both the industrial and academic community. Recently, backdoor attacks \cite{gu2019badnets,Trojannn,gao2020backdoor,gao2022imperceptible} have emerged as a new practical and stealthy threat to DNNs, for which the attacker plants pre-defined triggers to a small portion of the dataset and misleads the DNNs trained on such dataset to behave normally with benign inputs while classifying the input with trigger into the target class. To detect or mitigate the effect of backdoor, substantial efforts have been done in inversing triggers, splitting dataset, or pruning the DNNs, while fine-tuning, a natural choice for backdoor defense, has received much less attention. Although complex techniques such as unlearning and pruning have achieved remarkable performance, they usually come at the cost of accuracy on the original tasks. Additionally, the effectiveness of pruning is contingent upon the network structure, as highlighted by Wu \etal \cite{wu2021adversarial,wubackdoorbench}, underscoring the necessity for meticulous pruning strategies. In contrast, fine-tuning, a more general approach, can moderately restore the model's utility.

\begin{figure}[t]
\begin{center}
   \includegraphics[width=\linewidth]{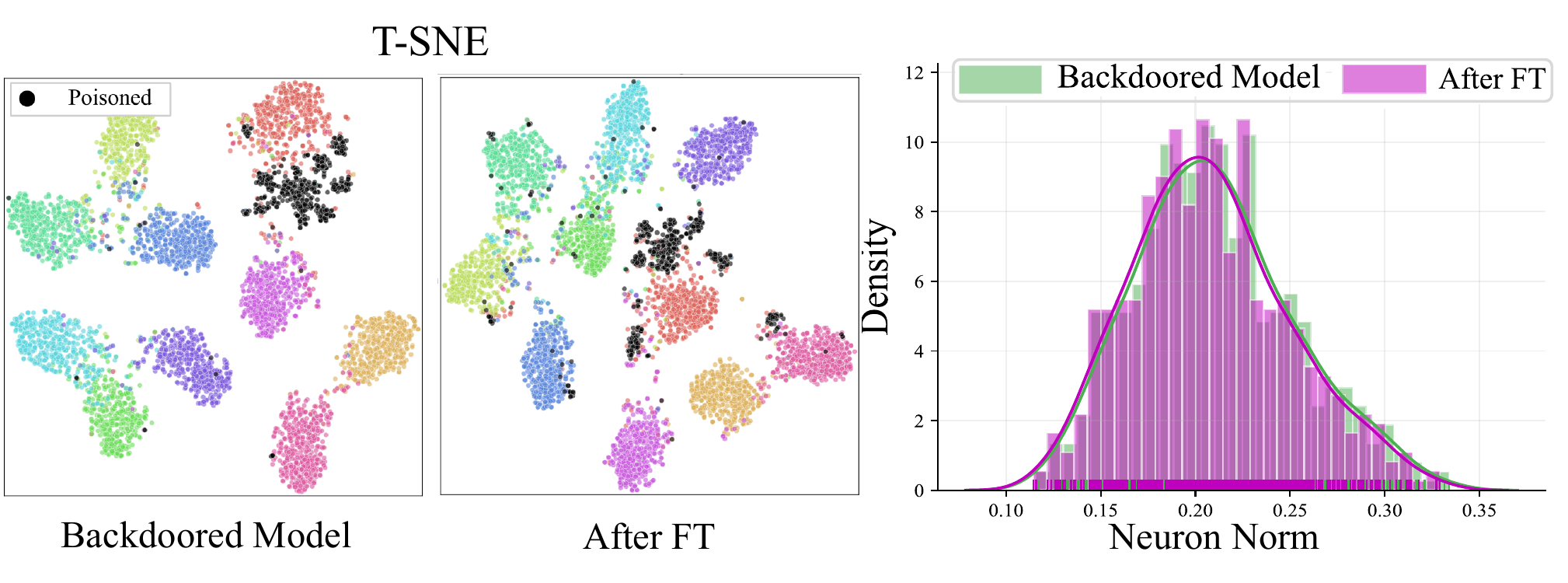}
\end{center}
   \vspace{-0.8em}
   \caption{Left: T-SNE \cite{van2008visualizing} visualization on the backdoored model and the model after fine-tuning. FT fails to remove backdoor effect. Right: the neuron weight norm distribution between the two model. The weight seems to have remained mostly unchanged after the fine-tuning process.}
   \vspace{-0.4cm}
\label{fig:tac}
\end{figure}

Although vanilla fine-tuning has been adopted as a component of some backdoor defense methods \cite{liu2018fine,li2021neural}, fine-tuning a backdoored model to remove the backdoor is still challenging when only limited benign data is given\cite{wubackdoorbench}. Previous work \cite{wubackdoorbench} has found that fine-tuning is a powerful technique in some situations, however, it cannot resist strong backdoor attacks such as Blended \cite{chen2017targeted} and LF \cite{zeng2021rethinking}. One of the possible reason is the backdoored model already fits the benign samples well; hence, vanilla fine-tuning can only make minor changes to the weights of neurons and fail to mitigate the backdoor effect, as demonstrated in Figure~\ref{fig:tac}. In this paper, we focus on the problem of designing a new objective function that can alter the backdoor-related weights and  help to remove the backdoor effect via fine-tuning.

To address this problem, we first take a closer look at the fine-tuning process from neurons' perspective. 
We empirically observe that the weight norm of neurons has a positive correlation with backdoor-related neurons in our experiment, which is also implied in \cite{zheng2022data}. Intuitively, the neurons with large norms can cause the backdoor features to override the normal features, making the model incorrectly pay attention to the trigger's feature. Motivated by the relationship between the neuron weight norms and the backdoor effect, we propose to adopt Sharpness-Aware Minimization (SAM) with adaptive perturbations \cite{foret2021sharpnessaware,zhuang2022surrogate} to fine-tune the backdoored model, which can revise the large outliers of weight norms and induce a more concentrated distribution of weight norms \cite{li2022efficient}. In detail, SAM considers a min-max formulation to encourage the weights in neighbors with an uniformly low loss. The adaptive constraints on perturbations can facilitate greater change of backdoor-related neurons. By leveraging SAM on the backdoored model, we empirically show that the model not only benefits from escaping the current local minima but also receives more perturbations on backdoor-related neurons than the normal weights. Therefore, SAM implicitly facilitates the learning of backdoored neurons and helps to remove the backdoor effect.

To demonstrate the effectiveness of our method, we conduct experiments on three benchmark datasets with two networks, and compare them to seven state-of-the-art defense methods. The results shows our method is competitive with and frequently superior to the best baseline. Our method is also robust across different components. Additionally, we empirically confirm that our strategy can take the place of fine-tuning, which can be used in conjunction with current backdoor defense techniques to make up for accuracy drop.

In summary, our main contributions are three-fold: \textbf{(1)} We reveal the reason of the weak backdoor defense performance of the vanilla fine-tuning based on a deep investigation from the perspective of backdoor-related neurons' weight changes.  \textbf{(2)} By leveraging SAM, we design an innovative fine-tuning paradigm to effectively remove the backdoor effect from a pre-trained backdoored model by perturbing the neurons.
\textbf{(3)} Experimental results and analyses demonstrate that the proposed method can achieve state-of-the-art performance among existing defense methods and boost existing defense methods based on fine-tuning. 

\comment{
In summary, our main contributions are three-fold: \textbf{(1)} We investigate the fine-tuning process of backdoored model and  establish the relation between neuron weight norm and backdoor effect. \textbf{(2)} By leveraging SAM, we design a new fine-tuning paradigm to remove the backdoor from a pre-trained model by perturbing the neurons with large norm.
\textbf{(3)} We empirically show that the proposed method can achieves state-of-the-art performance among existing defense methods and boost existing defense methods based on fine-tuning. }

%% file: contents/2_relatedwork.tex
\section{Related work}
\paragraph{Backdoor Attack.}
Several backdoor attacks have been proposed, including data poisoning attacks and training controllable attacks. In data poisoning attacks, BadNets \cite{gu2019badnets} is one of the most earliest attacks, in which they revise a small part of the data by patching a pre-defined pattern onto the images and relabeling them to the targeted class. Then the DNN trained on the poisoned dataset will be planted a backdoor. Blended \cite{chen2017targeted} designs a more strong backdoor attack by blending benign images with a whole pre-defined image. Recently, more advanced backdoors have been proposed to increase concealment of the triggers, such as LF \cite{zeng2021rethinking}, Wanet \cite{nguyen2021wanet}, and Input-aware \cite{nguyen2020input}. Training controllable backdoor attacks \cite{nguyen2021wanet, li2021invisible} assume that the attacker can control the training process, such that the attack can flexibly design triggers or decide the images to attack. To better evade backdoor detection, clean-label attacks\cite{shafahi2018poison,barni2019new} succeed by destroying the subject information of images and building a connection between the planted trigger and targeted label.

\vspace{-0.2cm}
\paragraph{Backdoor Defense.}
In general, backdoor defense methods can be categorized into two types: training-stage defenses and post-training defenses. Training-stage defenses \cite{li2021anti,chen2022effective} consider that a defender is given a backdoored dataset to train the model. The defender can leverage the different behaviors between benign and poisoned images in the training process to escape attacks, such as the loss dropping speed \cite{li2021anti} and clustering phenomenon in the feature space\cite{chen2019detecting,huang2022backdoor}. Most defense methods belong to post-training defenses \cite{wang2019neural, liu2018fine, zeng2022adversarial, wu2021adversarial, zheng2022data,zheng2022preactivation}, where the defender is given a suspicious model and has no access to the full training dataset. They need to remove backdoor threat by using a small set of benign samples. Post-training defenses can be roughly divided into fine-tuning based defenses (NC \cite{wang2019neural}, NAD \cite{li2021neural}, and i-BAU \cite{zeng2022adversarial} ) and pruning-based defenses ( FP \cite{liu2018fine} and ANP \cite{wu2021adversarial} ). FP \cite{liu2018fine} assume that poisoned and benign samples have different activation path. They remove backdoors by pruning the inactivated neurons of benign data and then fine-tuning the pruned model. ANP \cite{wu2021adversarial} assumes that backdoor-related neurons are more sensitive to adversarial neuron perturbations. They search for and mask these suspicious neurons by a minimax optimization on benign samples. NC \cite{wang2019neural} searches for a possible trigger by optimization and retrains the model by regularizing it to predict correctly on the images with the recovered trigger. NAD \cite{li2021neural} first fine-tunes a teacher model on a small subset of benign data and then fine-tunes a student model under the guidance of the teacher model. I-BAU \cite{zeng2022adversarial} borrows ideas from  universal adversarial perturbations and proposes the implicit backdoor adversarial unlearning algorithm to solve the minimax problem.

\vspace{-0.2cm}
\paragraph{Sharpness-Aware Minimization.}
Loss landscape has long been considered related to generalization in deep learning. Hochreiter and Schmidhuber \cite{hochreiter1997flat} have provided numerical support for the hypothesis that flat and wide minima generalize better than sharp minima. Chaudhari \etal \cite{chaudhari2019entropy} propose entropy-SGD to explicitly search for wider minima. Recently, SAM \cite{foret2021sharpnessaware} improves model generalization by simultaneously minimizing loss value and loss sharpness. Besides, several variants of SAM have been proposed \cite{zhuang2022surrogate,sunfedspeed,sun2023adasam,huangrobust,mi2022make,zhong2022improving} to search for flat minima. For example, ASAM \cite{kwon2021asam} proposes a new learning method for flat loss surface which is invariant to parameter re-scaling. GSAM \cite{zhuang2022surrogate} deﬁnes a surrogate gap and they minimize the surrogate gap and perturb loss synchronously. PGN \cite{yu2022robust} improves generalization by directly minimizing the loss function and the gradient norm. Randomed smoothing \cite{feng2020regularized} is another way to improve generalization and has been widely used in adversarial training \cite{cohen2019certified}. 

In contrast to existing post-processing defenses, our approach introduces a minimax formulation to enhance the process of fine-tuning for backdoor removal. Notably, our approach does not require any modification to the underlying network architecture, and preserves the model's utility. Furthermore, our work sheds light on the efficacy of SAM technique for fine-tuning in backdoor defenses.



%% file: contents/3_method.tex
\section{Methodology}

\subsection{Problem Formulation\label{sec3.1}}

\paragraph{Threat Model.}
We assume that an adversary carries out a backdoor attack on a DNN model $f_{\w}$ with weights $\w \in \mathbb{R}^d$, where $d$ is the number of parameters in the model.   The poisoning ratio is defined as the proportion of poisoned samples in the training dataset. The goal of the attacker is to make the model trained on the poisoned dataset classify the samples with triggers to the target labels while classifying clean samples normally. 

\paragraph{Defender's Goal.}
We consider that the defender is given a backdoored model and a few \textit{benign} samples $\Db$. The defender's goal is to fine-tune the model so that the benign data performance is maintained, and the backdoor effect is removed, \ie , the ratio of poisoned samples that are misclassified as the target label is low.

\subsection{Investigating the Vanilla Fine-Tuning\label{sec3.2}}
In the vanilla fine-tuning (FT) based backdoor defense \cite{liu2018fine}, it is assumed that limited benign samples $\Db$ (\eg, only $5\%$ benign samples), which are drawn from the same distribution as the original benign training dataset, are available to fine-tune the backdoored model. 
As evaluated in the latest backdoor learning benchmark, \ie, BackdoorBench \cite{wubackdoorbench}, FT has some effect on mitigating the backdoor behavior in some cases, but doesn't work well when facing several advanced backdoor attacks. 
We \textit{hypothesize} that since the backdoored model has already fitted the benign training samples $\Db$ well during the pre-training process, FT on $\Db$ cannot provide sufficient power to escape from the current solution (\ie, current model weights), such that the backdoor effect cannot be mitigated well.
 
\begin{figure}[h] 
\begin{center}
\includegraphics[width=\linewidth]{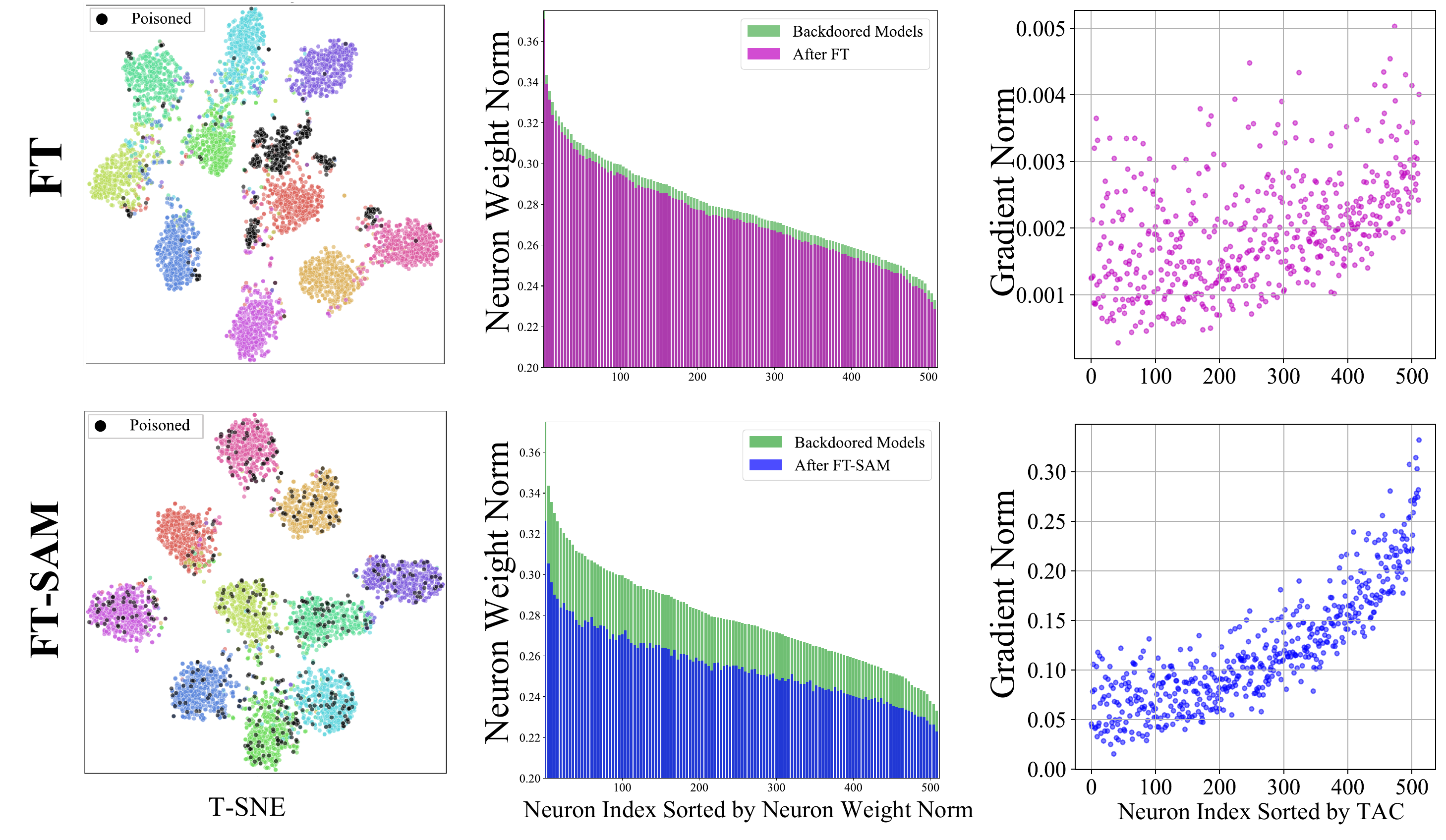}
\end{center}
\vspace{-0.8em}
   \caption{A comparison of the defense models by vanilla fine-tuning (FT, top row) and by FT-SAM (bottom row), respectively. The left column illustrates the T-SNE visualizations of the two models. The two figures in the middle depict the changes in neuron norms in the last convolution layer of the two models, sorted by the neuron weight index of the backdoored model. The last two figures present a comparison of the gradient norms for each neuron in the last convolution layer of the two models, which are calculated during the first batch of the first epoch.}
\label{method}
\end{figure}
To verify the above hypothesis, we conduct a deep investigation of the FT based backdoor defense. Specifically, we fine-tune the backdoored model using 5\% benign training samples for 100 epochs using the same learning rate as in the training of the backdoored model. All experiments adopt the Blended \cite{chen2017targeted} attack with poisoning ratio $10\%$, on CIFAR-10 dataset \cite{krizhevsky2009learning} and PreAct-ResNet18 model \cite{he2016identity}. 
The accuracy on benign testing dataset (\ie, benign accuracy) of this backdoored model is $93.44\%$, and the Attack Success rate (ASR) is $97.71\%$. After FT defense, the benign accuracy and ASR are changed to $92.48\%$ and $82.22\%$, respectively. 
As shown in the top row of Figure \ref{method}, we provide three consecutive perspectives to analyze the FT's effect in this experiment:
\begin{enumerate}
    \item \textbf{T-SNE Visualization.} The left-top T-SNE visualization in Figure \ref{method} shows the feature space of the fully connected layer in the FT defended model, on both benign testing images (\ie, colored points) and poisoned testing samples (\ie, black points). The poisoned samples are still clustered together. This explains its high ASR value. 
    \item \textbf{Changes of Neuron Weight Norms.} As shown in the middle-top sub-plot of Figure \ref{method}, we compare the changes of neuron weight norms in the last convolutional layer (containing 512 neurons) between the backdoored model (see green bins) and the FT defended model (see purple bins). It is observed that there are only slight changes on most neuron weights. It verifies that FT cannot give a new model that is far from the current model. 
    \item \textbf{Gradient Norms of Backdoor Related Neurons.} As shown in the right-top sub-plot of Figure \ref{method}, we further observe the gradient norms calculated on one mini-batch training data (\ie, 256 samples) during the FT process. We sort these 512 neurons by the TAC value \cite{zheng2022data} in ascending order.  
    TAC is proposed to measure the correlation between the backdoor effect and neurons. TAC in the $l^{th}$ layer is defined as the activation differences of channel-wise neurons in the $l^{th}$ layer between the benign samples and the corresponding poisoned ones in the model, and higher TAC value indicates stronger correlation.
    As shown in the figure, we can obtain two observations. \textbf{1)} All gradient norms are very small, which explains the slight neuron weight changes; \textbf{2)} There are not significant differences on gradient norms between right and left neurons. It implies that the weight changes between backdoor-related and non-backdoor-related neurons are similar. This explains why the backdoor effect is not mitigated well after FT.  
\end{enumerate}

\subsection{Proposed Method\label{sec3.3}}
\vspace{-0.2cm}
\paragraph{Min-Max Formulation.} Inspired by the findings in Section~\ref{sec3.2}, we now turn to design a strategy that can perturb the backdoor-related neurons. Motivated by the positive correlation between backdoor-related neurons and the neurons with large weight norms, we can naturally consider the adversarial perturbation on the weight of neurons. 
 We propose the following optimization problem:
\begin{equation}\label{eq:1}
    \min_{\w}  \max_{\boldsymbol{\|\T_{\w}^{-1}\epsilon\|_2\leq \rho}} \gL(\w+\boldsymbol{\epsilon}),
\end{equation}
where  $ \gL(\w + \boldsymbol{\epsilon}) = \mathbb{E}_{(\x,\y)\in \Db}\left[\ell(f_{\w+\boldsymbol{\epsilon}}(\x),\y)\right]$ with cross-entropy loss $\ell$, and $\rho > 0$ is the hyper-parameter for the budget of weight perturbation. Inspired by \cite{kwon2021asam}, we introduce $\T_{\w}=\text{diag}\left(|\w_1|,|\w_2|, \dots, |\w_d| \right) \in \mathbb{R}^{d\times d}$, where $\w_i$ is the $i$-th entry of $\w$, to set adaptive perturbation budget for different neurons and encourage larger perturbations to the neurons with lager weight norms, which are more likely to be related to the backdoor effect (Figure~\ref{fig4}). 

\begin{algorithm}[t]
\caption{Fine-tuning with SAM}\label{alg:ftsam}
\begin{algorithmic}[1]
\STATE \textbf{Input:} Training set $\Db$, backdoored model $f_{\w}$, learning rate $\eta>0$, perturbation bound $\rho>0$, loss function $\gL$, max iteration number $T$.
\STATE \textbf{Output:} Model fine-tuned with SAM.
\STATE Initialize $\w_0$.
\FOR {$t=0,...,T-1$}
\STATE Sample a mini-batch $\mathcal{B}$ from $\Db$;
\STATE Update $\T_{\w_t}$; 
\STATE Update $\hat{\boldsymbol{\epsilon}}_{t+1}$ via Equation \ref{eq:2} \wrt $\mathcal{B}$;
\STATE Update weights: $\boldsymbol{w}_{t+1}=\boldsymbol{w}_t-\eta \nabla_{\w}\gL(\w_{t}+\beps_{t+1})$ \wrt~$\mathcal{B}$; 
\ENDFOR
\RETURN $f_{\w_T}$ 
\end{algorithmic}
\end{algorithm}

\vspace{-0.2cm}
\paragraph{Optimization.} As described in Algorithm~\ref{alg:ftsam}, Problem~(\ref{eq:1}) can be efficiently solved by alternatively updating $\w$ and $\beps$, as follows:

\noindent
\textbf{Inner Maximization:} Given model weight $\w_{t}$, the weight perturbation $\beps$ could be updated by solving the following sub-problem:
\begin{equation}\label{eq:sub1}
    \max_{\boldsymbol{\|\T_{\w_{t}}^{-1}\epsilon\|_2\leq \rho}} \gL(\w_t+\boldsymbol{\epsilon}).
\end{equation}
According to Taylor expansion, the first-order approximation of the solution to Problem~(\ref{eq:sub1}) is
    \begin{equation}
    \label{eq:2}
    \begin{aligned}
    	\beps_{t+1} &= \arg \max_{\boldsymbol{\|\T_{\w_{t}}^{-1}\epsilon\|_2\leq \rho}} \gL(\w_{t}+ \boldsymbol{\epsilon}) \\
    	& \approx \arg \max_{\boldsymbol{\|\T_{\w_{t}}^{-1}\epsilon\|_2\leq \rho}} \gL(\w_{t}) + \boldsymbol{\epsilon}^\top\nabla_{\w} \gL(\w_{t})\\
    	&= \rho \frac{\T_{\w_{t}}^2\nabla_{\w} \gL(\w_{t})}{\left\|\T_{\w_{t}}\nabla_{\w} \gL(\w_{t})\right\|_2}.
    \end{aligned}
    \end{equation}
Due to the space limit, the detailed derivation of the above update will be provided in Section \ref{app:A} of \textbf{Appendix}.

\noindent
\textbf{Outer Minimization:} Given $\beps_{t+1}$, the model weight $\w$ can be updated by solving the following sub-problem
\begin{equation}
    \label{eq:sub2}
    \min_{\w} \gL(\w+\beps_{t+1}),
\end{equation}
which can be optimized by stochastic gradient descent, i.e., $\w_{t+1} = \w_t - \eta \nabla_{\w}\gL(\w_t+\beps_{t+1})$ where $\eta$ is the learning rate. Here, we update $\w_t$ and $\beps_t$ once for each batch of data.

\noindent
\textbf{Remark.} The bottom row of Figure \ref{method} shows the effectiveness of FT-SAM on backdoor defenses. The T-SNE visualization in the left-bottom shows the poisoned features are dispersed and lie closely to features of benign samples. The middle-bottom sub-plot shows the neurons with large norms are severely perturbed following our protection strategy compared to FT. Moreover, we visualize the gradient calculated by FT-SAM in Figure \ref{method}. It can be observed that the norms of the gradient have a positive correlation with backdoor related neurons. This explains how our method outperforms vanilla FT. We also analyze the norm distribution in each channels of the defense model. The figure is presented in Section \ref{sec4.4}. It shows that the model searches a solution with more balanced neurons compared to backdoored model. As a result, the high utilization of the network parameters prevents network decisions from being dominated by individual features, especially backdoor features.

%% file: contents/4_experiment.tex
\section{Experiment}

\subsection{Experimental Setup\label{sec4.1}}
\input{tables/table1}
\input{tables/table2}

\paragraph{Attack Settings.} We consider 10 popular state-of-the-art (SOTA) backdoor attacks: BadNets \cite{gu2019badnets} with two attack settings (BadNets-A2O and BadNets-A2A refer to attacking one class and all classes, respectively) , blended backdoor attack (Blended) \cite{chen2017targeted}, Input-aware dynamic backdoor attack (Input-aware)\cite{nguyen2020input}, Clean-label attack (CLA)\cite{shafahi2018poison}, Low frequency attack (LF) \cite{zeng2021rethinking}, Sinusoidal signal backdoor attack (SIG) \cite{barni2019new}, Sample-specific backdoor attack (SSBA) \cite{li2021invisible}, Trojan backdoor attack (Trojan) \cite{Trojannn}, and Warping-based poisoned networks (WaNet) \cite{nguyen2021wanet}. We follow the default attack conﬁguration as in BackdoorBench \cite{wubackdoorbench} for a fair comparison, such as the trigger patterns and optimization hyper-parameters. The poisoning ratio is set to 10\% in all attacks and the $0^{th}$ label is set to be the targeted label except for BadNets-A2A, in which  the target labels for original labels $y$ are set to $y_t=(y+1) \mod{C}$. Here $C$ is total number of classes and $\mod$ is short for "modulus". We evaluate all the attacks on 3 benchmark datasets, CIFAR-10 \cite{krizhevsky2009learning}, Tiny ImageNet \cite{le2015tiny}, and GTSRB \cite{stallkamp2011german} over two networks, PreAct-ResNet18 \cite{he2016identity} and VGG19-BN \cite{simonyan2014very} except for two clean-label attacks SIG and CLA, where the $10\%$ poisoning ratio cannot be reached by attacking only one class. To test the robustness of our method, we also compare our method to SOTA methods with a $5\%$ poisoning ratio on CIFAR-10 and Tiny ImageNet on PreAct-ResNet18. Due to the space limit, more implementations details about backdoor attacks and the comparison results with $5\%$ poisoning ratio can be found in Section \ref{app:B} and \ref{app:C} of \textbf{Appendix}.


\vspace{-0.2cm}
\paragraph{Defense Settings.}
We compare the proposed method with vanilla fine-tuning (FT) and seven SOTA backdoor defense methods: Fine-pruning (FP) \cite{liu2018fine}, NAD \cite{li2021neural}, AC \cite{chen2019detecting}, NC \cite{wang2019neural}, ANP \cite{wu2021adversarial}, ABL \cite{li2021anti}, and i-BAU \cite{zeng2022adversarial}. All the defense methods can access 5\% benign training data except for AC and ABL, which use the entire poisoned dataset and train a model from scratch. We follow the default configurations for SOTA defense as in BackdoorBench \cite{wubackdoorbench}. We use a learning rate of $0.01$ with batch size 256 for 100 epochs on CIFAR-10 and Tiny ImageNet, and 50 epochs on GTSRB for FT and FT-SAM. The analysis of sensitivity to different number of benign training samples can be found in Section \ref{sec4.3}. For FT-SAM, the most crucial hyper-parameter is the perturbation radius $\rho$. We set $\rho=2$ for CIFAR-10 and $\rho=8$ for Tiny ImageNet and GTSRB on PreAct-ResNet18. Due to the space limit, more details about defense settings can be found in Section \ref{app:B} of \textbf{Appendix}.

\input{tables/table3}

\vspace{-0.2cm}
\paragraph{Evaluation Metric.} We use three metrics to evaluate the performance of different defenses: ACCuracy on benign data ($\textbf{ACC}$), Attack Success Rate ($\textbf{ASR}$), and Defense Effectiveness Rating ($\textbf{DER}$). ASR measures the proportion of backdoor samples that are successfully misclassified to the target label. DER $\in [0,1]$ is firstly proposed in this work to evaluate defense performance considering both ACC and ASR. It is defined as follows:
\begin{equation}
    \text{DER} = [\max(0, \Delta \text{ASR})-\max(0,\Delta \text{ACC})+1]/2,
\end{equation}
where $\Delta \text{ASR}$ denotes the drop in ASR after applying defense, and $\Delta \text{ACC}$ represents the drop in ACC after applying defense. For instance, a value of $\text{DER}=1$ means the defense successfully reduces the ASR from $1$ to $0$ without any drop in ACC; $\text{DER}=0$ means ACC drops from $1$ to $0$ and ASR doesn't change. The $\max$ is added to the metric since the increase of ACC or ASR rarely occurs in defenses. A superior defense is indicated by a lower ASR, higher ACC, and higher DER. To ensure a fair comparison between different strategies for the target label, we remove samples whose ground-truth labels already belong to the target class. \textbf{Note} that among all defenses, the one with the best performance is indicated in \textbf{boldface} and the value with \underline{underline} denotes the second-best result.

\begin{figure*}[!t]
\begin{center}
\includegraphics[width=\linewidth]{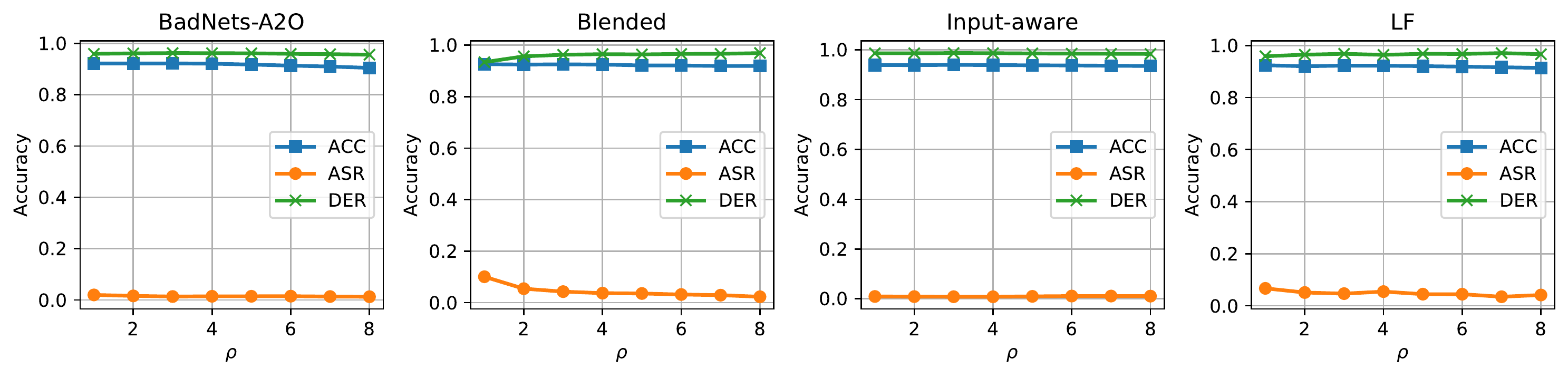}
\end{center}
\vspace{-0.5cm}
\caption{Performance of FT-SAM with $\rho$ from $1$ to $8$ against different attacks on CIFAR-10 and 5\% poisoning ratio with PreAct-ResNet18.}
\label{fig:rho}
\end{figure*}

\subsection{Experimental Results\label{sec4.2}}
We verify the effectiveness of our method by comparing it against the seven SOTA defense methods on CIFAR-10 and Tiny ImageNet with 10\% poisoning ratio on PreAct-ResNet18. The results are presented in Table \ref{table1} and Table \ref{table2}. As shown in Table \ref{table1}, Badnets-A2O and Wanet can be defended by almost all the defense methods. FT shows promising defense performance and maintains ACC on several attacks, but it cannot resist complex attacks, such as Blended, LF and SSBA. The results of NAD are very similar to FT as both methods fine-tune the model with limited data. I-BAU demonstrates a noticeable effect against almost all attacks with average ASR $<6\%$, but it sacrifices ACC to achieve a robust model, as evidenced by a low DER. ANP and ABL also show potential in defending against some attacks but their results are unstable, with fluctuating ASR , low ACC and low DER on different attacks. The sensitivity of the pruning threshold among different attacks in ANP may explain this result, while ABL's process of combining learning and unlearning may harm the model's utility. NC performs comparably well in some attacks while the average DER is low, indicating that NC's resilience is not that high. 
In comparison, our approach receives a high DER in nost cases, indicating the effectiveness of our method in defending against various attacks. It demonstrates power to decrease ASR on average (2.47\%) across all attacks. 

Table \ref{table2} presents the experimental results on Tiny ImageNet with PreAct-ResNet18. We observe that all compared defense methods fail to maintain both ACC and ASR on complex attacks, which is reflected in an low DER. FT, FP, and NAD demonstrate similar defense performance as they cannot defend against complex attacks. ABL is successful in removing backdoors while reducing ACC synchronously, while i-BAU fails on Tiny ImageNet, possibly due to the larger input size which increases the difficulty of minimax optimization. In contrast, the proposed method shows robustness against all the attacks, with only a slight drop in ACC and remarkably high DER. The defense results on the GTSRB dataset and the performance on VGG19-BN can be found in Section \ref{app:C} of \textbf{Appendix}.

\subsection{Ablation Studies\label{sec4.3}}
\input{tables/table4}

\begin{figure*}[t]
\begin{center}
\includegraphics[width=\linewidth]{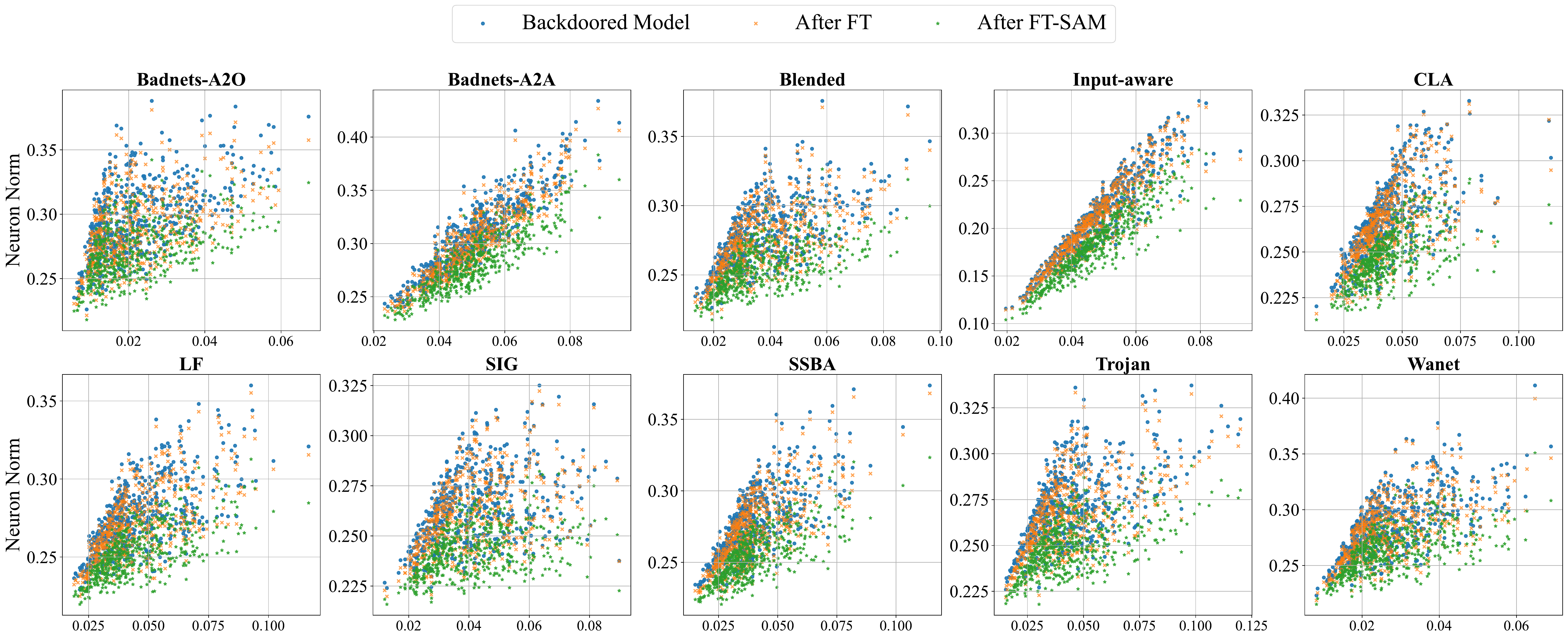}
\end{center}
\vspace{-0.2cm}
\caption{Neuron weight norm comparison to the backdoored models \wrt TAC between FT and FT-SAM defenses against different attacks on CIFAR-10 and 5\% poisoning ratio with PreAct-ResNet18.}
\label{fig4}
\vspace{-0.4cm}
\end{figure*}

\vspace{-0.2cm}
\paragraph{Performance with Different Values of Hyper-parameter~$\rho$.} 
The most crucial hyper-parameter in our defense approach is the constraint bound $\rho$ imposed on the perturbation $\epsilon$. A higher value of $\rho$ increases the weight perturbation, thereby improving the network's robustness. However, in cases where we are given limited training data, a smaller value of $\rho$ can help maintain the model's performance while reducing the effectiveness of defense. Here we evaluate the sensitivity of $\rho$ by conducting four complex attacks using a learning rate of 0.01 and different value of $\rho$. Figure \ref{fig:rho} displays the defense results. A smaller value of $\rho$ may not completely remove backdoors, especially for complex attacks. But it shows that FT-SAM can enhance the model's robustness and exhibit a certain level of ACC and DER when faced with different $\rho$. Overall, the hyper-parameter $\rho$ is not very sensitive, and a wide range of values can be selected without significantly impacting the model's performance.

\vspace{-0.2cm}
\paragraph{Performance under Different Components.}
To evaluate the effectiveness of FT-SAM in various scenarios, we conducted experiments with different numbers of benign training samples, backbones, and poisoning ratios. Table \ref{table3} presents the defense results of FT-SAM on the CIFAR-10 dataset using PreAct-ResNet18 with a $10\%$ poisoning ratio under different ratios of benign samples. The hyper-parameter $\rho$ is set to $2$ across all experiments. We observed that FT-SAM demonstrates a robust defense mechanism across various numbers of benign samples, with only a modest decrease in performance given $1\%$ benign samples. Contrarily, different attacks cause different trends in the effectiveness of FT at various number of benign samples, and poor results can be observed through the exceptionally low DER especially when the ratio is low. In contrast, our method exhibits consistently high DER. Further results on the performance of FT-SAM with different backbones (VGG19-BN) and poisoning ratios can be found in Section \ref{app:C} of \textbf{Appendix}.

\subsection{Further Analysis\label{sec4.4}}
\paragraph{Combination with SOTA Defenses.} 
So far, we have demonstrated the effectiveness of our method in backdoor removal. As discussed in Section \ref{sec3.3}, FT-SAM, as a kind of fine-tuning method, shows superiority over vanilla fine-tuning and has the potential to replace it in defense processes. Moreover, we hypothesize that FT-SAM can also enhance pruning-based defense methods, which suffer from performance drops if the defense configuration is not well optimized. To verify our hypothesis and demonstrate the versatility of FT-SAM, we combined it with two existing post-processing defense methods: FP \cite{liu2018fine} and ANP \cite{wu2021adversarial}. FP first prunes the suspicious neurons of the model and then fine-tunes the pruned model with limited samples. We replace fine-tuning to FT-SAM in the second step of FP. ANP identifies the backdoor neurons that mostly enlarge the loss function and then masks these neurons. We keep the mask computed by ANP and fine-tune it with FT-SAM. The experiment is conducted on the CIFAR-10 with PreAct-ResNet18. We also display results for pure pruning, as well as the combination of ANP and fine-tuning (\textit{ANP + FT}) for a fair comparison. The results are shown in Table \ref{table4}. The original defense methods showed susceptibility to various attacks, including ANP against Blended and FP against CLA. Additionally, ANP shows a low ACC and DER. Although the \textit{ANP + FT} sometimes worked, it performs poorly in other attacks. On average, our proposed approach improves both defense strategies with a high DER. This result may inspire the development of new robust defense strategies with the help of FT-SAM.

\begin{figure}[t]
\vspace{-0.3cm}
\begin{center}
\includegraphics[width=\linewidth]{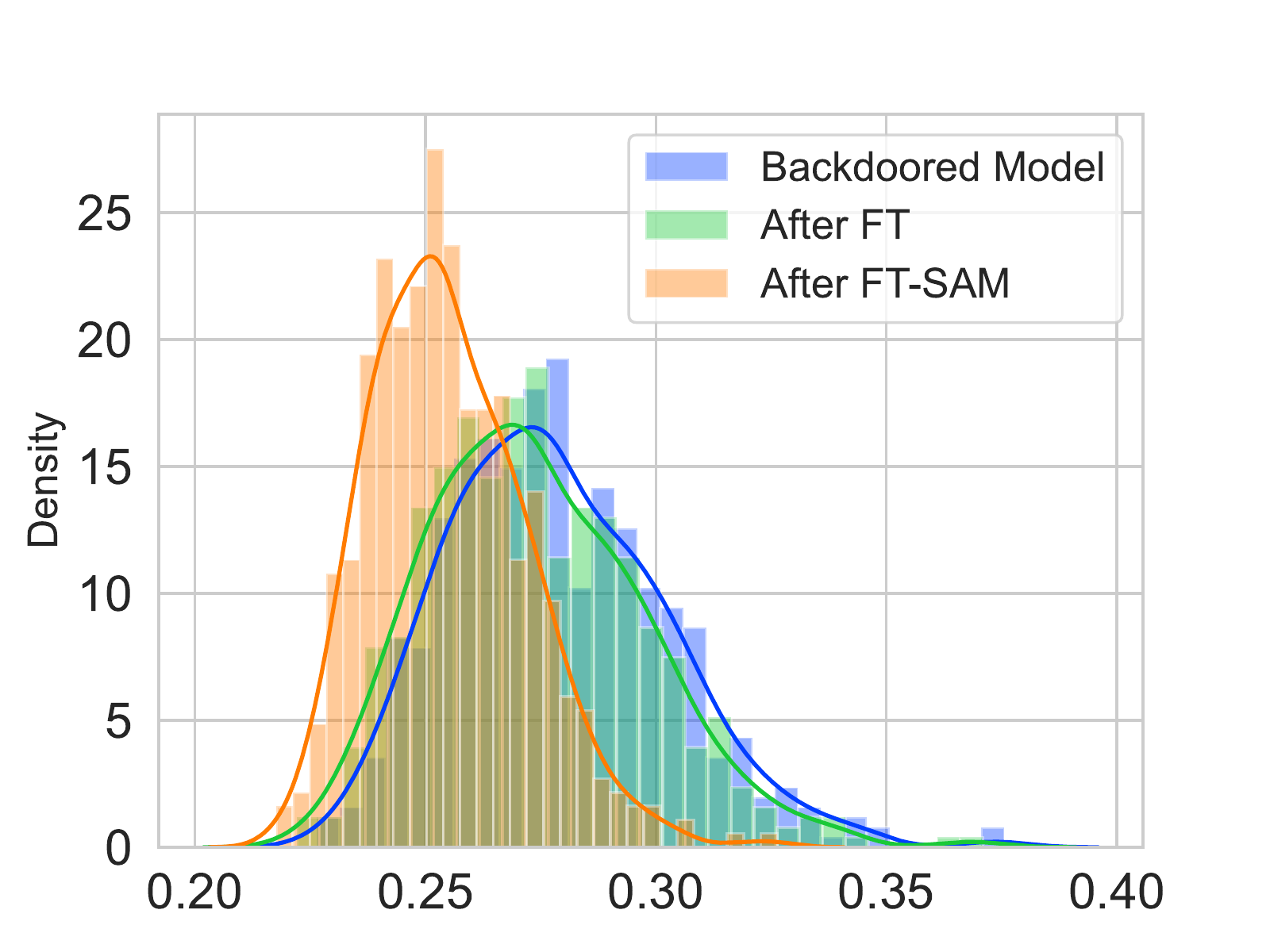}
\end{center}
\vspace{-0.4cm}
\caption{Parameters distribution comparison to the backdoored model between FT and FT-SAM defenses on CIFAR-10 and 5\% poisoning ratio with PreAct-ResNet18.}
\label{fig5}
\vspace{-0.2cm}
\end{figure}

\vspace{-0.2cm}
\paragraph{Analysis of Weight Changes and Weight Norm Distribution of FT-SAM.} 
We analyze our defense method from neuron norm's perspective. A scatter plot chart of neuron weight norm \wrt TAC of ten attacks is presented in Figure \ref{fig4}. It is observed that the neurons weight norm and TAC are highly correlated. The neuron weight norm of the models experience only minor changes before and after FT. However, after applying FT-SAM, there is a notable decrease in the overall neuron weight norm. Moreover, the neurons with higher norms experience a greater change. These results suggest that our method effectively perturbs the neurons that are linked to the backdoor. Furthermore, as shown in Figure \ref{fig5}, our approach uniformly depresses the norm for all neurons. Therefore, the network relies more evenly on neurons for decision-making, leading to a more trustworthy model.

\begin{figure}[t]
\begin{center}
\includegraphics[width=\linewidth]{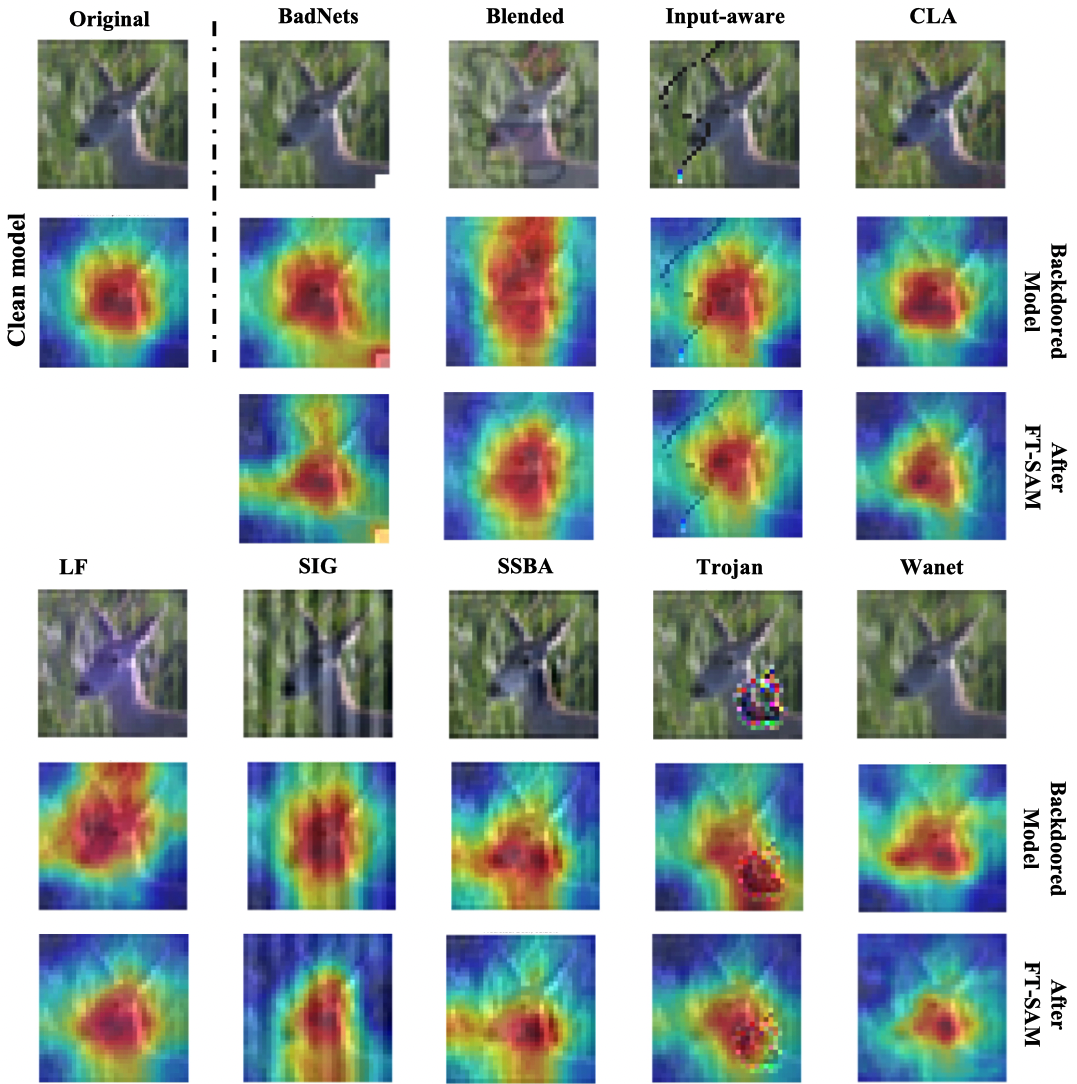}
\end{center}
\vspace{-0.2cm}
   \caption{Grad-CAM \cite{selvaraju2017grad} visualization of regions contributed to model decision under different attacks by FT-SAM defense comparing to the backdoored models on CIFAR-10 dataset and 5\% poisoning ratio with PreAct-ResNet18.}
\label{fig6}
\vspace{-0.2cm}
\end{figure}

\vspace{-0.3cm}
\paragraph{Grad-CAM Visualization under Different Attacks.} 
Grad-CAM \cite{selvaraju2017grad} figures can provide insights into how a neural network makes its predictions. If the original model generates a strong signal in a subject region of the image that is highly relevant for the classification task, then this could indicate that the defense mechanism has successfully removed the backdoor. Figure \ref{fig6} displays the benign image of a deer and its Grad-CAM figure, along with the samples from nine attacks and their Grad-CAM figures. As shown in the figure, compared to the backdoored model, all the Grad-CAM figures of the defense models focus on the subject region of the image, \ie, the head of deer instead of the triggers. This demonstrates that the backdoor has been eliminated successfully.



%% file: tables/table1.tex
\begin{table*}[!t]
\caption{Comparison with the state-of-the-art defenses on \textbf{CIFAR-10} dataset with 5\% benign data on PreAct-ResNet18 (\%).}
\centering
\renewcommand\arraystretch{1.5} 
\resizebox{\textwidth}{!}{%
\begin{tabular}{@{}c|c|c|c|c|c|c|c|c|c|c@{}}
\toprule
\multirow{2}{*}{Attack} &
  \textbf{Backdoored} &
  \textbf{FT} &
  \textbf{FP} \cite{liu2018fine} &
  \textbf{NAD} \cite{li2021neural} &
  \textbf{AC} \cite{chen2019detecting} &
  \textbf{NC} \cite{wang2019neural} &
  \textbf{ANP} \cite{wu2021adversarial} &
  \textbf{ABL} \cite{li2021anti} &
  \textbf{i-BAU} \cite{zeng2022adversarial} &
  \textbf{FT-SAM(Ours)} \\
 &
  ACC/ASR &
  ACC/ASR/DER &
  ACC/ASR/DER &
  ACC/ASR/DER &
  ACC/ASR/DER &
  ACC/ASR/DER &
  ACC/ASR/DER &
  ACC/ASR/DER &
  ACC/ASR/DER &
  ACC/ASR/DER \\ \hline \hline
BadNets-A2O\cite{gu2019badnets} &
  91.82/93.79 &
  90.29/1.70/95.28 &
  \underline{91.77}/\underline{0.84}/\textbf{96.45} &
  88.82/1.96/94.42 &
  48.84/16.57/67.12 &
  57.22/0.90/79.14 &
  91.65/3.83/94.89 &
  80.10/\textbf{0.00}/91.03 &
  87.43/4.48/92.46 &
  \textbf{92.21}/1.63/\underline{96.08} \\
BadNets-A2A\cite{gu2019badnets} &
  91.89/74.42 &
  91.07/1.16/86.22 &
  \underline{92.05}/1.31/\underline{86.56} &
  90.73/1.61/85.83 &
  87.23/67.03/51.37 &
  89.79/\underline{1.11}/85.61 &
  \textbf{92.33}/2.56/85.93 &
  44.39/40.65/43.14 &
  89.39/1.29/85.32 &
  91.87/\textbf{1.03}/\textbf{86.69} \\
Blended\cite{chen2017targeted} &
  93.44/97.71 &
  92.48/82.22/57.26 &
  \underline{92.57}/8.32/\underline{94.26} &
  92.09/55.04/70.66 &
  88.82/95.10/49.00 &
  91.91/84.31/55.94 &
  \textbf{93.00}/57.38/69.95 &
  74.31/\textbf{0.10}/89.24 &
  88.24/6.00/93.26 &
  92.44/\underline{4.91}/\textbf{95.90} \\
Input-aware\cite{nguyen2020input} &
  94.03/98.35 &
  93.00/65.85/65.74 &
  94.05/10.95/93.70 &
  \underline{94.08}/10.43/\underline{93.96} &
  51.37/90.94/32.38 &
  \textbf{94.11}/98.98/50.00 &
  94.06/11.10/93.63 &
  50.58/98.82/28.28 &
  89.91/\underline{8.92}/92.66 &
  93.76/\textbf{1.07}/\textbf{98.51} \\
CLA\cite{shafahi2018poison} &
  84.55/99.93 &
  90.38/10.76/94.59 &
  90.67/78.72/60.61 &
  90.01/8.53/95.70 &
  81.57/99.11/48.92 &
  \textbf{90.87}/4.56/97.69 &
  82.55/\underline{0.18}/\textbf{98.88} &
  68.14/\textbf{0.00}/91.76 &
  85.66/18.99/90.47 &
  \underline{90.72}/3.52/\textbf{98.21} \\
LF\cite{zeng2021rethinking} &
  93.01/99.06 &
  92.37/93.89/52.26 &
  92.05/21.32/88.39 &
  91.72/75.47/61.15 &
  52.28/94.34/31.99 &
  \textbf{93.01}/99.06/50.00 &
  \underline{92.53}/26.38/86.10 &
  71.68/\textbf{0.86}/88.44 &
  88.92/11.99/\underline{91.49} &
  91.07/\underline{3.81}/\textbf{96.65} \\
SIG\cite{barni2019new} &
  84.49/97.87 &
  90.47/5.74/96.06 &
  \underline{90.81}/7.06/95.41 &
  90.05/6.60/95.63 &
  81.33/98.23/48.42 &
  84.50/97.87/50.00 &
  83.87/97.24/50.00 &
  48.06/\textbf{0.00}/80.72 &
  85.87/1.32/\underline{98.27} &
  \textbf{91.16}/\underline{0.80}/\textbf{98.53} \\
SSBA\cite{li2021invisible} &
  92.88/97.07 &
  \underline{92.47}/90.04/53.31 &
  92.21/20.27/88.07 &
  92.15/70.77/62.79 &
  46.75/67.63/41.65 &
  \textbf{92.88}/97.07/50.00 &
  92.02/16.18/90.01 &
  79.87/\textbf{0.33}/91.86 &
  86.53/2.89/\underline{93.91} &
  92.12/\underline{2.80}/\textbf{96.75} \\
Trojan\cite{Trojannn} &
  93.47/99.99 &
  92.59/35.50/81.80 &
  92.24/67.73/65.51 &
  92.18/5.77/96.47 &
  89.47/100.00/48.00 &
  91.85/51.03/73.67 &
  \underline{92.71}/84.82/57.20 &
  70.70/\textbf{0.02}/88.60 &
  89.29/\underline{0.54}/\textbf{97.63} &
  \textbf{92.75}/4.12/\underline{97.57} \\
Wanet\cite{nguyen2021wanet} &
  92.80/98.90 &
  \underline{93.14}/1.26/98.82 &
  92.94/\textbf{0.66}/\textbf{99.12} &
  93.07/\underline{0.73}/\underline{99.08} &
  52.81/11.86/73.52 &
  92.80/98.90/50.00 &
  \textbf{93.24}/1.54/98.68 &
  67.23/92.97/40.18 &
  90.70/0.88/97.96 &
  92.87/0.96/98.97 \\ \midrule
Avg &
  91.24/95.71 &
  91.83/38.81/78.45 &
  \textbf{92.14}/21.72/86.99 &
  91.49/23.69/86.01 &
  68.05/74.08/49.22 &
  87.89/63.38/64.49 &
  90.80/30.12/82.57 &
  65.51/23.37/73.30 &
  88.19/\underline{5.73}/\underline{93.47} &
  \underline{92.10}/\textbf{2.47}/\textbf{96.62} \\ \bottomrule
\end{tabular}%
\label{table1}
}
\end{table*}

%% file: tables/table2.tex
\begin{table*}[!t]
\caption{Comparison with the state-of-the-art defenses on \textbf{Tiny ImageNet} dataset with 5\% benign data on PreAct-ResNet18 (\%).}
\centering
\renewcommand\arraystretch{1.5} 
\resizebox{\textwidth}{!}{%
\begin{tabular}{@{}c|c|c|c|c|c|c|c|c|c|c@{}}
\toprule
\multirow{2}{*}{Attack} &
  \textbf{Backdoored} &
  \textbf{FT} &
  \textbf{FP} \cite{liu2018fine} &
  \textbf{NAD} \cite{li2021neural} &
  \textbf{AC} \cite{chen2019detecting} &
  \textbf{NC} \cite{wang2019neural} &
  \textbf{ANP} \cite{wu2021adversarial} &
  \textbf{ABL} \cite{li2021anti} &
  \textbf{i-BAU} \cite{zeng2022adversarial} &
  \textbf{FT-SAM(Ours)} \\
 &
  ACC/ASR &
  ACC/ASR/DER &
  ACC/ASR/DER &
  ACC/ASR/DER &
  ACC/ASR/DER &
  ACC/ASR/DER &
  ACC/ASR/DER &
  ACC/ASR/DER &
  ACC/ASR/DER &
  ACC/ASR/DER \\ \hline \hline
BadNets-A2O\cite{gu2019badnets} &
  56.12/99.90 &
  \underline{55.56}/0.44/\textbf{99.45} &
  48.81/0.66/95.96 &
  48.35/0.27/95.93 &
  49.21/99.76/46.62 &
  \textbf{56.12}/99.90/50.00 &
  47.34/\textbf{0.00}/95.56 &
  48.34/\textbf{0.00}/96.06 &
  51.63/95.92/49.74 &
  51.91/0.21/\underline{97.74} \\
BadNets-A2A\cite{gu2019badnets} &
  55.99/27.81 &
  \textbf{55.04}/22.28/52.29 &
  47.88/3.19/58.26 &
  48.29/\underline{2.30}/\underline{58.91} &
  47.71/13.15/53.19 &
  \underline{54.12}/18.72/53.61 &
  40.70/2.39/55.07 &
  49.60/29.44/46.81 &
  53.52/12.89/56.23 &
  52.24/\textbf{2.09}/\textbf{60.99} \\
Blended\cite{chen2017targeted} &
  55.53/97.57 &
  \textbf{54.74}/87.18/54.80 &
  47.45/34.40/77.54 &
  49.52/67.60/61.98 &
  48.51/96.50/47.02 &
  \underline{52.79}/\textbf{0.04}/\textbf{97.39} &
  40.21/28.78/76.73 &
  47.95/\underline{0.10}/94.94 &
  49.30/26.34/82.50 &
  50.81/1.03/\underline{95.91} \\
Input-aware\cite{nguyen2020input} &
  57.67/99.19 &
  \textbf{57.86}/0.68/\textbf{99.26} &
  49.18/3.75/93.48 &
  50.08/0.61/95.50 &
  49.48/98.73/46.14 &
  \underline{56.15}/84.64/56.52 &
  50.62/\underline{0.46}/95.84 &
  49.42/\textbf{0.10}/95.42 &
  53.96/1.29/\underline{97.10} &
  52.69/1.01/96.60 \\
LF\cite{zeng2021rethinking} &
  55.21/98.51 &
  \textbf{54.53}/94.14/51.85 &
  48.18/63.83/63.83 &
  49.61/58.01/67.45 &
  49.68/98.17/47.41 &
  53.08/90.48/52.95 &
  41.75/65.98/59.54 &
  45.37/\textbf{0.02}/\underline{94.33} &
  \underline{53.65}/94.27/51.34 &
  51.30/\underline{3.58}/\textbf{95.51} \\
SSBA\cite{li2021invisible} &
  55.97/97.69 &
  \textbf{55.17}/92.08/52.40 &
  48.06/52.25/68.76 &
  47.67/69.47/59.96 &
  49.02/97.44/46.65 &
  \underline{53.30}/\underline{0.26}/\textbf{97.38} &
  41.83/14.24/84.65 &
  47.39/\textbf{0.00}/94.55 &
  52.39/84.64/54.73 &
  51.87/0.38/\underline{96.60} \\
Trojan\cite{Trojannn} &
  56.48/99.97 &
  \textbf{55.70}/37.11/81.04 &
  45.96/8.88/90.28 &
  48.83/1.01/95.66 &
  49.82/99.96/46.68 &
  \underline{54.43}/1.54/\textbf{98.19} &
  45.36/0.53/94.16 &
  46.31/\textbf{0.00}/94.90 &
  51.85/99.15/48.10 &
  52.28/\underline{0.21}/\underline{97.78} \\
Wanet\cite{nguyen2021wanet} &
  57.81/96.50 &
  \underline{57.37}/0.18/\textbf{97.94} &
  50.35/1.37/93.83 &
  50.02/0.87/93.92 &
  48.99/99.68/45.59 &
  \textbf{57.81}/96.50/50.00 &
  30.34/\textbf{0.00}/84.51 &
  47.01/\underline{0.02}/92.84 &
  53.04/69.82/60.95 &
  54.32/0.79/\underline{96.11} \\ \midrule
Avg &
  56.35/89.64 &
  \textbf{55.75}/41.76/73.64 &
  48.23/21.04/80.24 &
  49.05/25.02/78.66 &
  49.05/87.92/47.21 &
  \underline{54.73}/49.01/69.50 &
  42.27/14.05/80.76 &
  47.67/\underline{3.71}/\underline{88.63} &
  52.42/60.54/62.59 &
  52.18/\textbf{1.16}/\textbf{92.16} \\ \bottomrule
\end{tabular}%
\label{table2}
}
\end{table*}

%% file: tables/table3.tex
\begin{table*}[!t]
\caption{Performance with different benign ratio under different attacks on CIFAR-10 dataset with 10\% poisoning ratio on PreAct-ResNet18 (\%).}

\centering
\renewcommand\arraystretch{1.2} 
\resizebox{\textwidth}{!}{%
\begin{tabular}{@{}c|c|c|c|c|c|c|c|c|c|c|c@{}}
\toprule
\multirow{2}{*}{Benign Ratio} &
  Model &
  BadNets-A2O\cite{gu2019badnets} &
  BadNets-A2A\cite{gu2019badnets} &
  Blended\cite{chen2017targeted} &
  Input-aware\cite{nguyen2020input} &
  CLA\cite{shafahi2018poison} &
  LF\cite{zeng2021rethinking} &
  SIG\cite{barni2019new} &
  SSBA\cite{li2021invisible} &
  Trojan\cite{Trojannn}	&	
  Wanet\cite{nguyen2021wanet} \\
 &
   &
  ACC/ASR/DER &
  ACC/ASR/DER &
  ACC/ASR/DER &
  ACC/ASR/DER &
  ACC/ASR/DER &
  ACC/ASR/DER &
  ACC/ASR/DER &
  ACC/ASR/DER &
  ACC/ASR/DER &
  ACC/ASR/DER \\ \hline\hline
 &
  Backdoored &
  91.82/93.79/- &
  91.89/74.42/- &
  93.44/97.71/- &
  94.03/98.35/- &
  84.55/99.93/- &
  93.01/99.06/- &
  92.65/95.89/- &
  84.49/97.87/- &	
  92.88/97.07/-	&	
  93.47/99.99/- \\ \midrule
\multirow{2}{*}{10\%} &
  FT &
  91.67/\textbf{1.17}/96.24 &
  90.42/1.61/85.67 &
  \textbf{92.62}/77.20/59.85 &
  94.17/10.44/93.96 &
  \textbf{91.53}/5.68/97.13 &
  92.12/69.22/64.47 &
  91.18/\textbf{0.88}/96.77 &	
  \textbf{92.42}/65.32/66.27	&
  92.68/99.61/49.90 &	\textbf{93.57}/1.50/99.24\\
 &
  \textbf{FT-SAM} &
  \textbf{91.94}/1.26/\textbf{96.27} &
  \textbf{92.46}/\textbf{1.01}/\textbf{86.71} &
  92.53/\textbf{3.94}/\textbf{96.43} &
  \textbf{94.22}/\textbf{0.93}/\textbf{98.71} &
  91.44/\textbf{4.90}/\textbf{97.52} &
  \textbf{92.64}/\textbf{3.83}/\textbf{97.43} &
  \textbf{91.46}/1.13/\textbf{96.78} &	
  91.75/\textbf{2.63}/\textbf{97.62}	&	
  \textbf{92.94}/\textbf{2.50}/\textbf{97.28}&	93.23/\textbf{0.78}/\textbf{99.48} \\ \midrule
\multirow{2}{*}{5\%} &
  FT &
  90.29/1.70/95.28 &
  91.07/1.16/86.22 &
  \textbf{92.48}/82.22/57.26 &
  93.00/65.85/65.74 &
  90.38/10.76/94.59 &
  \textbf{92.37}/93.89/52.26 &
  90.47/5.74/93.98 &	
  \textbf{92.47}/90.04/53.91	&	
  92.59/35.50/80.64 &	\textbf{93.14}/1.26/99.20 \\
 &
  \textbf{FT-SAM} &
  \textbf{92.21}/\textbf{1.63}/\textbf{96.08} &
  \textbf{91.87}/\textbf{1.03}/\textbf{86.69} &
  92.44/\textbf{4.91}/\textbf{95.90} &
  \textbf{93.76}/\textbf{1.07}/\textbf{98.51} &
  \textbf{90.72}/\textbf{3.52}/\textbf{98.21} &
  91.07/\textbf{3.81}/\textbf{96.65} &
  \textbf{91.16}/\textbf{0.80}/\textbf{96.80} &	
  92.12/\textbf{2.80}/\textbf{97.53}	&	
  \textbf{92.75}/\textbf{4.12}/\textbf{96.41} &	92.87/\textbf{0.96}/\textbf{99.22} \\ \midrule
\multirow{2}{*}{1\%} &
  FT &
  \textbf{89.25}/6.14/92.54 &
  \textbf{91.98}/1.42/\textbf{86.50} &
  \textbf{92.09}/87.98/54.19 &
  92.23/69.51/63.52 &
  88.12/11.61/94.16 &
  \textbf{92.36}/98.87/49.77 &
  87.80/3.07/93.99 &	
  \textbf{92.08}/94.87/51.50	&	
  \textbf{92.61}/99.70/49.87 &	\textbf{92.65}/9.79/94.69 \\
 &
  \textbf{FT-SAM} &
  88.96/\textbf{1.29}/\textbf{94.82} &
  90.43/\textbf{1.12}/85.92 &
  90.83/\textbf{3.01}/\textbf{96.05} &
  \textbf{93.12}/\textbf{0.81}/\textbf{98.31} &
  \textbf{88.74}/\textbf{3.13}/\textbf{98.40} &
  90.88/\textbf{4.64}/\textbf{96.14} &
  \textbf{88.31}/\textbf{1.00}/\textbf{95.27} &	
  91.11/\textbf{1.51}/\textbf{98.18}	&	
  90.63/\textbf{4.36}/\textbf{95.23} &	
  90.79/\textbf{0.91}/\textbf{98.20} \\ \bottomrule
\end{tabular}%
\label{table3}
}

\end{table*}

%% file: tables/table4.tex
\begin{table*}[!t]
\centering
\renewcommand\arraystretch{1.2}  
\caption{Combination with SOTA defenses on CIFAR-10 dataset with 5\% benign data on PreAct-ResNet18 (\%).}

\resizebox{\textwidth}{!}{%
\begin{tabular}{@{}c|c|c|c|c|c|c|c|c@{}}
\toprule
\multirow{2}{*}{Attack} &
  BadNets-A2O\cite{gu2019badnets} &
  Blended\cite{chen2017targeted} &
  Input-aware\cite{nguyen2020input} &
  CLA\cite{shafahi2018poison} &
  LF\cite{zeng2021rethinking} &
  SIG\cite{barni2019new} &
  SSBA\cite{li2021invisible} &
  Wanet\cite{nguyen2021wanet} \\
 &
  ACC/ASR/DER &
  ACC/ASR/DER &
  ACC/ASR/DER &
  ACC/ASR/DER &
  ACC/ASR/DER &
  ACC/ASR/DER &
  ACC/ASR/DER &
  ACC/ASR/DER \\ \hline \hline
 Backdoored &
  91.82/93.79/- &
  93.44/97.71/- &
  94.03/98.35/- &
  84.55/99.93/- &
  93.01/99.06/- &
  92.65/95.89/- &
  84.49/97.87/- &
  93.47/99.99/- \\ \midrule
Pruning &
  82.52/97.22/45.35 &
  81.25/99.31/43.91 &
  84.66/99.90/45.32 &
  75.42/99.72/45.54 &
  83.22/99.78/45.11 &
  75.57/78.57/50.12 &
  80.75/98.53/48.13 &
  83.38/99.84/45.03 \\ 
Pruning+FT(FP \cite{liu2018fine}) &
  \underline{91.77}/\underline{0.84}/\textbf{96.45} &
  92.57/8.32/94.26 &
  94.05/10.95/93.70 &
  90.67/78.72/60.61 &
  92.05/21.32/88.39 &
  \underline{90.81}/7.06/93.50 &
  \underline{92.21}/20.27/88.80 &
  92.94/\underline{0.66}/\underline{99.40} \\ 
Pruning + \textbf{FT-SAM} &
  91.20/\textbf{0.62}/\underline{96.27} &
  92.07/\underline{5.42}/\underline{95.46} &
  93.84/1.14/\underline{98.51} &
  90.18/33.76/83.09 &
  92.03/\underline{17.42}/\underline{90.33} &
  90.30/5.06/94.24 &
  91.83/\underline{14.73}/\underline{91.57} &
  92.52/\textbf{0.57}/99.24 \\ \midrule
ANP \cite{wu2021adversarial} &
  91.65/3.83/94.89 &
  \textbf{93.00}/57.38/69.95 &
  \underline{94.06}/11.10/93.63 &
  82.55/\textbf{0.18}/\underline{98.88} &
  \underline{92.53}/26.38/86.10 &
  83.87/97.24/45.61 &
  92.02/16.18/90.84 &
  \underline{93.24}/1.54/99.11 \\ 
ANP \cite{wu2021adversarial} + FT &
  \textbf{92.24}/1.41/96.19 &
  \underline{92.90}/42.28/77.45 &
  \textbf{94.17}/\underline{1.11}/\textbf{98.62} &
  \textbf{91.47}/6.44/96.74 &
  \textbf{92.71}/63.33/67.71 &
  \textbf{91.22}/\textbf{0.08}/\textbf{97.19} &
  \textbf{92.57}/35.46/81.21 &
  \textbf{93.36}/\underline{0.66}/\textbf{99.61} \\ 
ANP \cite{wu2021adversarial} + \textbf{FT-SAM} &
  90.99/1.12/95.92 &
  91.51/\textbf{2.57}/\textbf{96.61} &
  93.03/\textbf{1.09}/98.13 &
  \underline{91.08}/\underline{2.09}/\textbf{98.92} &
  91.71/\textbf{4.00}/\textbf{96.88} &
  89.57/\textbf{0.08}/\underline{96.37} &
  91.49/\textbf{4.16}/\textbf{96.86} &
  91.90/0.78/98.82 \\ \bottomrule
\end{tabular}%
\label{table4}
}
\vspace{-0.2cm}
\end{table*}

%% file: contents/5_conclusion.tex
\section{Conclusion}
In this work, we investigate the impact of fine-tuning on backdoor defenses and provide insights into why fine-tuning fails from a neuron-level perspective. Specifically, we explore the relationship between the norm of network neurons and their contribution to backdoor attacks, and find that neurons with larger norms contribute more to backdoor attacks. Leveraging this observation, we propose a novel fine-tuning technique, dubbed FT-SAM, that employs sharpness-aware minimization to perturb backdoor-related neurons. We empirically demonstrate that our method can significantly reduce the weight norm of the backdoor-related neurons and shows its effectiveness by investigating the gradient of neuron weight computed by FT-SAM. Extensive experiments demonstrate that our method reliably eliminates the injected backdoor and offers the highest robustness against various cutting-edge backdoor attacks while preserving high accuracy. Finally, integrating our method with other defense methods demonstrates FT-SAM is a promising defense strategy against backdoor attacks.

%% file: contents/appendix.tex
\appendix
\section*{Appendix}

\section{More Algorithmic Details on The Proposed Method\label{app:A}}
We provide a detailed derivation of the optimization problem in Section 3 of the main script here. The constraint optimization problem is defined as follows:
\begin{equation}\label{eq:1a}
    \min_{\w}  \max_{\boldsymbol{\|\T_{\w}^{-1}\epsilon\|_2\leq \rho}} \gL(\w+\boldsymbol{\epsilon}),
\end{equation}
where  $ \gL(\w + \boldsymbol{\epsilon}) = \mathbb{E}_{(\x,\y)\in \Db}\left[\ell(f_{\w+\boldsymbol{\epsilon}}(\x),\y)\right]$ with cross-entropy loss $\ell$, $\rho > 0$ is the hyper-parameter for the budget of weight perturbation, and $\T$ is a diagonal matrix. 

\paragraph{Optimization.} Problem~(\ref{eq:1a}) can be efficiently solved by alternatively updating $\w$ and $\beps$, as follows:

\noindent
\textbf{Inner Maximization:} Given model weight $\w_{t}$, the weight perturbation $\beps$ could be updated by solving the following sub-problem:
\begin{equation}\label{eq:sub1a}
    \max_{\boldsymbol{\|\T_{\w_{t}}^{-1}\epsilon\|_2\leq \rho}} \gL(\w_t+\boldsymbol{\epsilon}).
\end{equation}
Define $\tilde{\boldsymbol{\epsilon}}=\T_{\w}^{-1} \boldsymbol{\epsilon}$. According to first-order Taylor expansion, the approximation of the solution to Problem~(\ref{eq:sub1a}) is
    \begin{equation}
    \label{eq:2a}
    \begin{aligned}
    	\tilde{\beps}_{t+1} &= \arg \max_{\|\tilde{\beps}\|_2\leq \rho} \gL(\w_{t}+  \T_{\w_t} \tilde{\beps}) \\
    	& \approx \arg \max_{\boldsymbol{\|\T_{\w_{t}}^{-1}\epsilon\|_2\leq \rho}} \gL(\w_{t}) + \tilde{\beps}^\top\T_{\w_t}\nabla_{\w} \gL(\w_{t})\\
     &= \arg \max_{\boldsymbol{\|\T_{\w_{t}}^{-1}\epsilon\|_2\leq \rho}} \tilde{\beps}^\top\T_{\w_t}\nabla_{\w} \gL(\w_{t})\\
    &=	\rho \frac{\T_{\w_{t}}\nabla_{\w} \gL(\w_{t})} {\left\|\T_{\w_{t}}\nabla_{\w} \gL(\w_{t})\right\|_2}. 
    \end{aligned}
    \end{equation}
Thus the inner problem can be solved as:
\begin{equation}
    \beps_{t+1} = \T_{\w_t} \tilde{\beps}_{t+1} = \rho \frac{\T_{\w_{t}}^2\nabla_{\w} \gL(\w_{t})}{\left\|\T_{\w_{t}}\nabla_{\w} \gL(\w_{t})\right\|_2}.
\end{equation}

\noindent
\textbf{Outer Minimization:} Given $\beps_{t+1}$, the model weight $\w$ can be updated by solving the following sub-problem:
\begin{equation}
    \label{eq:sub3a}
    \min_{\w} \gL(\w+\beps_{t+1}),
\end{equation}
which can be optimized by stochastic gradient descent, i.e., $\w_{t+1} = \w_t - \eta \nabla_{\w}\gL(\w_t+\beps_{t+1})$ where $\eta$ is the learning rate.

\section{More Implementation Details\label{app:B}}
\paragraph{Datasets.} We evaluate our method on CIFAR-10 \cite{krizhevsky2009learning}, Tiny ImageNet \cite{le2015tiny}, and GTSRB \cite{stallkamp2011german} following the benchmarks \cite{wubackdoorbench}. For details, CIFAR-10 contains 60,000 images from 10 classes, with 5000 images per class for training and 1000 images per class for testing. Each image has a size of $32 \times 32$. Tiny ImageNet is a subset of ImageNet, which contains 100,000 training samples and 10,000 testing samples over 200 classes. Each image has a size of $64 \times 64$. GTSRB contains 39209 and 12630 images for training and testing from 43 classes. Each image has a size of $32 \times 32$.
\paragraph{Models.}
We evaluate our method on PreAct-ResNet18 \cite{he2016identity} and VGG19-BN \cite{simonyan2014very} networks. We compare our method with SOTA defense methods on three datasets and the two networks with a $10\%$ poisining ratio and $5\%$ clean samples for defense. To study the effectiveness of our method under different poisoning ratios, we compare with SOTA defense methods on CIFAR-10 dataset and PreAct-ResNet18 network on $5\%$ and $1\%$ poisoning ratios.

\paragraph{Attack Details.} We introduce some details about the backdoor attacks here. For BadNets-A2O and BadNets-A2A \cite{gu2019badnets}, we patch a $3 \times 3$ white square in the lower right corner of the images for CIFAR-10 and GTSRB datasets, and $6 \times 6$ white square for Tiny ImageNet. For Blended \cite{chen2017targeted}, we blend the poisoned samples with a Hello-Ketty image and the blended ratio is $0.1$. 

\paragraph{Defense Details.} The seven SOTA defense methods can be divided into two types based on what the defender is given. AC \cite{chen2019detecting} and ABL \cite{li2021anti} assumes that the defender is given a poisoned dataset, while the remaining six defense methods assumes that the defender can acquire a subset of clean samples and a backdoored model. The learning rate for all methods is set to $0.01$, and the batch size is set to $256$. The threshold for ANP \cite{wu2021adversarial} is set to $0.4$ since we find that the recommended threshold $0.2$ fails to remove backdoors. For FT, the training epochs is set to 100 for CIFAR-10 and Tiny ImageNet, and 50 for GTSRB dataset. All other settings are consistent with those in BackdoorBench \cite{wubackdoorbench}. 

\paragraph{Details of Proposed Method.}
The most crucial hyper-parameter in FT-SAM is the perturbation radius $\rho$. We set $\rho=2$ for CIFAR-10 and $\rho=8$ for Tiny ImageNet and GTSRB on PreAct-ResNet18. For VGG19-BN, $\rho$ is set to 6 for all the three datasets. The epochs is set to 100 for CIFAR-10 and Tiny ImageNet, and 50 for GTSRB dataset. When the adaptive perturbation $\T$ is not applied to $\w$, the perturbation budget should be small to maintain the clean accuracy, where it is set to $0.5$ in this work. All the experiments are conducted with using SGD with momentum 0.9 and weight decay $1\mathrm{e}{-4}$.

\section{Defense Results in Comparison to SOTA Defenses\label{app:C}}
The defense performances of our method compared to the seven SOTA defense methods on GTSRB with PreAct-ResNet18 network and on the three datasets with VGG19-BN networks are displayed in Table \ref{tablea1} to Table \ref{tablea4}. \textbf{Note} that among all defenses, the one with the best performance is indicated in \textbf{boldface} and the value with \underline{underline} denotes the second-best result.

As shown in these tables, all the defense methods fails to balance the performance on both the clean accuracy (ACC) and the attack success rate (ASR) in all the situations except for FT-SAM, which is robust across all the attacks, datasets and backbones. The average defense effectiveness rating (DER) and ASR rank first among these defenses. Although the ACC of proposed method has dropped slightly, it usually does not fall below 1\% on average.

\input{tables_appendix/table1} 
\input{tables_appendix/table2}
\input{tables_appendix/table3}
\input{tables_appendix/table4}

\section{Defense Results under Different Poisoning Ratios  \label{app:D}}
To show the robustness of our method, we also test the defense performance when the poisoning ratios are $5\%$ and $1\%$, respectively. The defense results on CIFAR10 dataset with $5\%$ benign data on PreAct-ResNet18 are shown in Table \ref{tablea5} and Table \ref{tablea6}. As shown in the tables, attack performance drops when the poisoning ratio is only $1\%$ except for Trojan, which demonstrates the power of Trojan attack. FT-SAM significantly outperforms the other defense methods especially when the poisoning ratio is only $1\%$. 

\section{Ablation Study of The Effectiveness of $\T_{\w}$ \label{app:E}}
In this section, we first study the effectiveness of the adaptive constraint $\T_{\w}$, then we show the experimental results on the defense that directly regularizes the $l_2$ weight norm.
\paragraph{Effectiveness of The Adaptive Constraint $\T_{\w}$.}
The constraint without adaptive perturbation is equal to the situation where $\T_{\w}$ is set to identity matrix. The comparison result is shown in Table \ref{tablea9}. As shown in the table, the method without adaptive constraints has a lower ACC and a higher ASR on average. This gap is more pronounced when encountering complex attacks. It demonstrates the necessity of the adaptive constraint to the perturbation in FT-SAM. 
\paragraph{Defense results of The $l_2$ weight norm regularization.} To show the effectiveness of FT-SAM, we also test the defense performance by directly finetuning with regularizing the $l_2$ norm on the network parameters, \ie, the loss function is
\begin{equation}\label{eq:l2_norm}
    \min_{\w} \mathbb{E}_{(\x,\y)\in \Db}\left[\ell(f_{\w}(\x),\y)\right] + \gamma \|\w\|^2_2,
\end{equation}
where $\gamma>0$ is the hyper-parameter. We test this method on several complex attacks and the results under different values of $\gamma$ are shown in Table \ref{tablea10}. It is observed that regularizing weights can also weaken backdoor attacks to a certain extent. However, the hyper-parameter $\gamma$ is very sensitive to different attacks, and removing backdoors completely usually results in a large drop in clean accuracy. On the contrary, our method is more robust to different attacks, showing the effectiveness of our method on perturbing the backdoor-related weights.

\input{tables_appendix/table5}
\input{tables_appendix/table6}
\input{tables_appendix/table9}
\input{tables_appendix/table10}

\section{Visualization Analysis \label{app:F}}
\begin{figure*}[]
\begin{center}
   \includegraphics[width=0.8\linewidth]{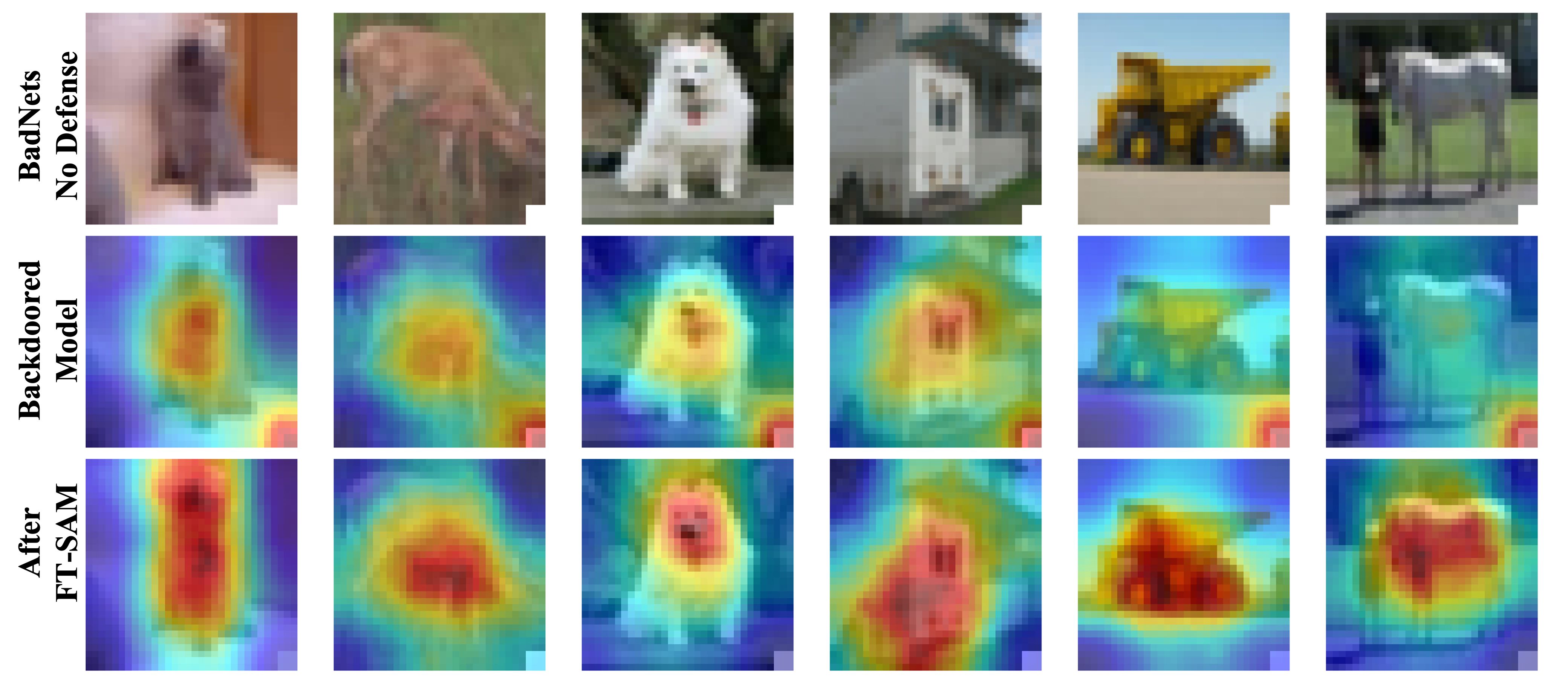}
\end{center}
   \vspace{-0.8em}
   \caption{Grad-CAM visualization of regions contributed to model decision under BadNets attack and FT-SAM defense with PreAct-ResNet18 on CIFAR-10.}
   \vspace{-0.4cm}
\label{cam:badnet}
\end{figure*}

\begin{figure*}[]
\begin{center}
   \includegraphics[width=0.8\linewidth]{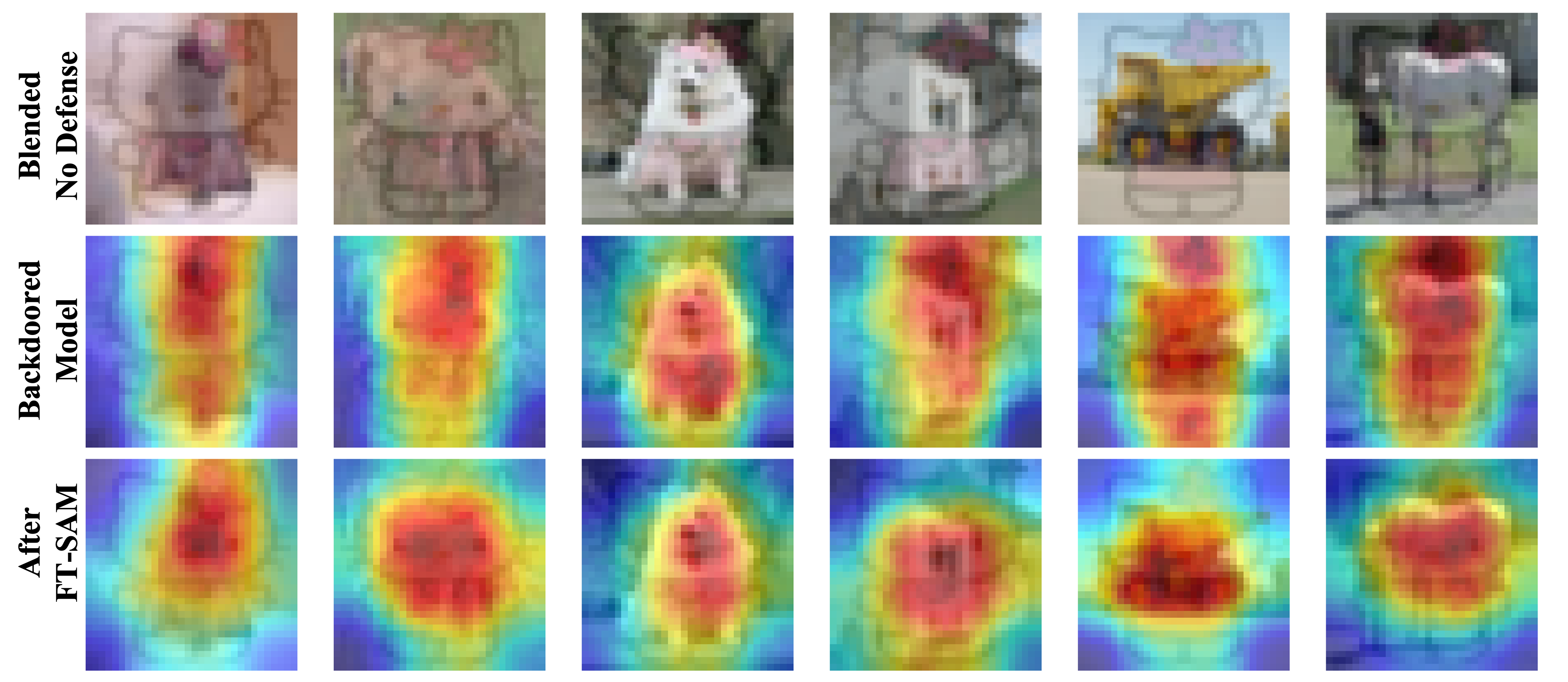}
\end{center}
   \vspace{-0.8em}
   \caption{Grad-CAM visualization of regions contributed to model decision under Blended attack and FT-SAM defense with PreAct-ResNet18 on CIFAR-10.}
   \vspace{-0.4cm}
\label{cam:blended}
\end{figure*}

\begin{figure*}[]
\begin{center}
   \includegraphics[width=0.8\linewidth]{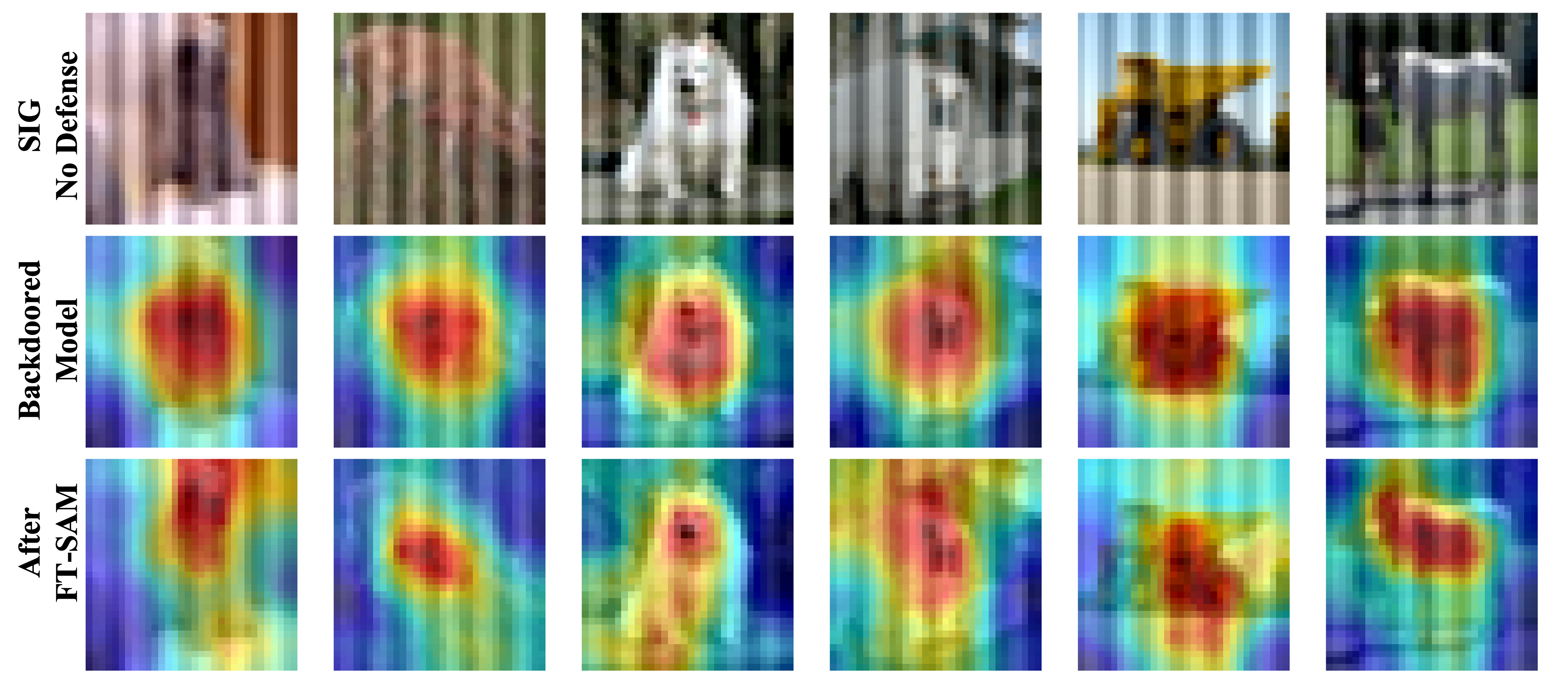}
\end{center}
   \vspace{-0.8em}
   \caption{Grad-CAM visualization of regions contributed to model decision under SIG attack and FT-SAM defense with PreAct-ResNet18 on CIFAR-10.}
   \vspace{-0.4cm}
\label{cam:sig}
\end{figure*}

\begin{figure*}[]
\begin{center}
   \includegraphics[width=0.8\linewidth]{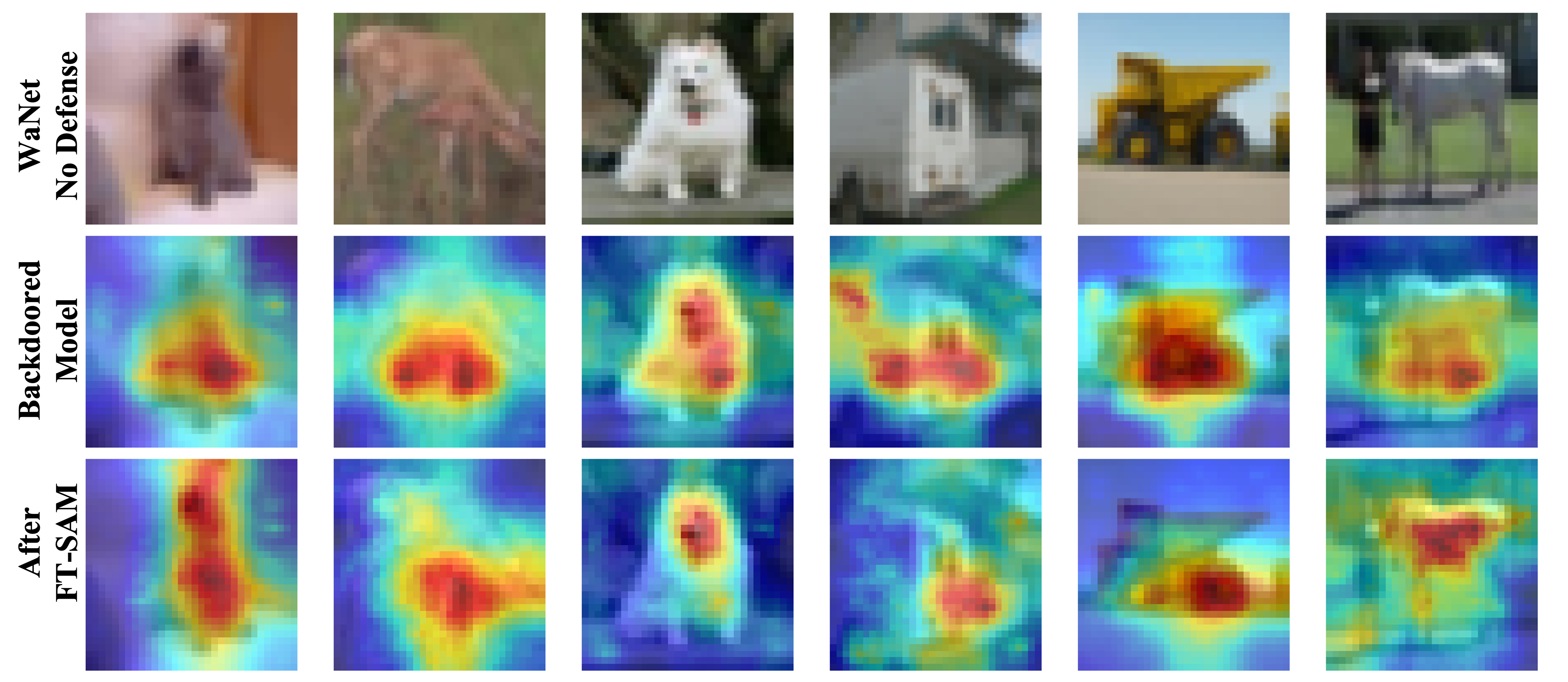}
\end{center}
   \vspace{-0.8em}
   \caption{Grad-CAM visualization of regions contributed to model decision under Wanet attack and FT-SAM defense with PreAct-ResNet18 on CIFAR-10.}
   \vspace{-0.4cm}
\label{cam:wanet}
\end{figure*}

\paragraph{Grad-CAM Visualization.} Figure \ref{cam:badnet} to \ref{cam:wanet} show the defense effect of our method on BadNets, Blended, SIG, and Wanet attacks by Grad-CAM \cite{selvaraju2017grad}. The top rows show the poisoned samples, while the second and third rows show the Grad-CAM figures on the backdoored models and the defense models, respectively. Figure \ref{cam:badnet} to \ref{cam:sig} belong to visible backdoor attacks. Comparing the highlighted area of the heat maps of the backdoored models and defense models, the defense models concentrate on the subject region of the images instead of the trigger features. Figure \ref{cam:wanet} shows the invisible backdoor attack. The defense models focus more on the subject region, where the backdoored models show similar areas of interest in all these images.

\begin{figure*}[]
\begin{center}
   \includegraphics[width=0.9\linewidth]{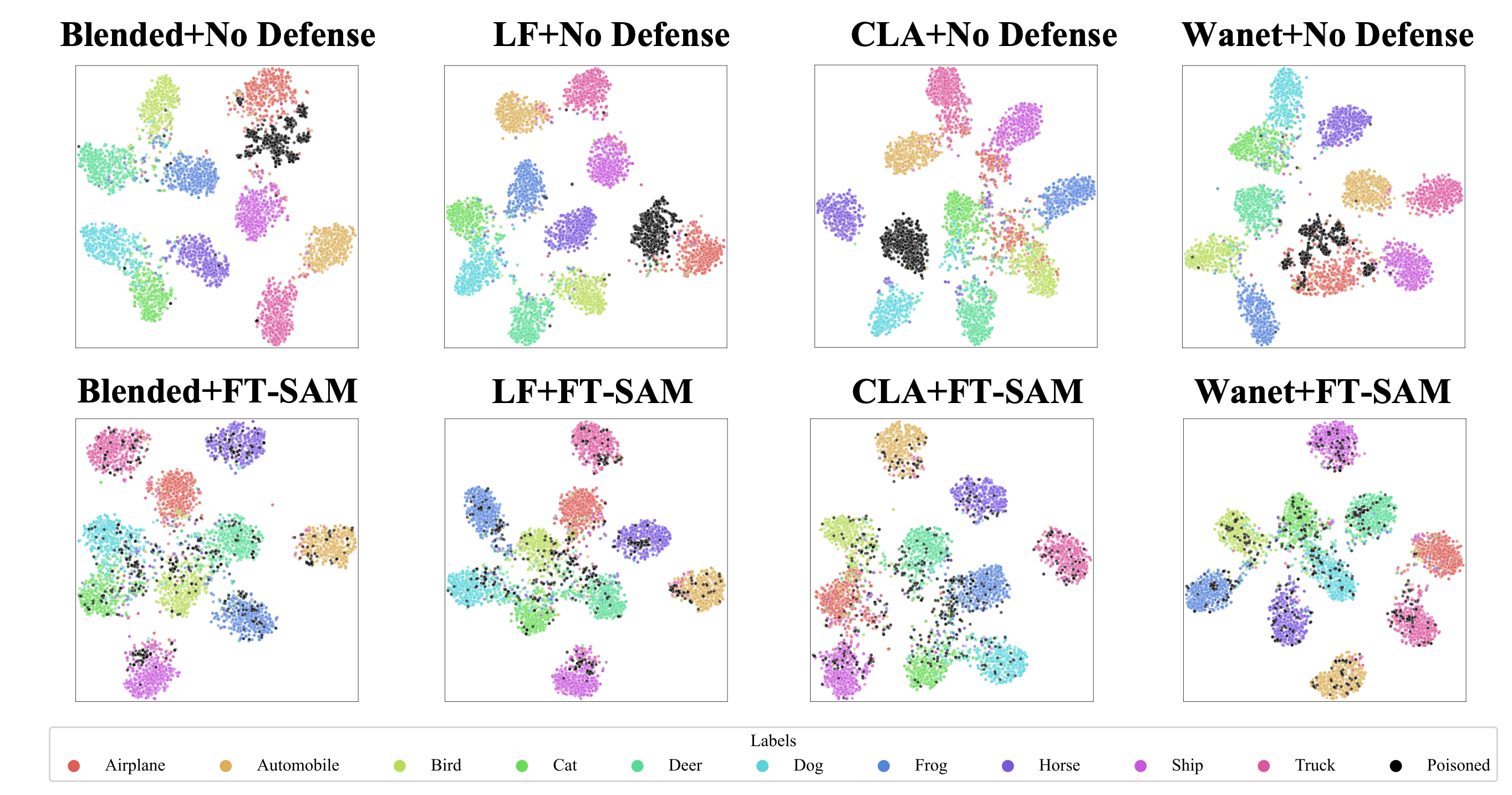}
\end{center}
   \vspace{-0.8em}
   \caption{T-SNE visualization under different backdoor attacks and FT-SAM defense models with PreAct-ResNet18 on CIFAR-10.}
   \vspace{-0.4cm}
\label{append:tsne}
\end{figure*}

\paragraph{T-SNE Visualization.} We provide more T-SNE \cite{van2008visualizing} visualization figures of our method as shown in Figure \ref{append:tsne}. compared to the first row which exhibits clustering of poisoned features in the feature space of the backdoored models, the proposed defense method successfully break up these poisoned features and make them distribute around the normal features.

\clearpage

%% file: tables_appendix/table1.tex
\begin{table*}[!t]

\caption{Comparison with the state-of-the-art defenses on \textbf{GTSRB} dataset with 5\% benign data on PreAct-ResNet18 (\%).}
\centering
\renewcommand\arraystretch{1.6} 
\resizebox{\textwidth}{!}{%
\begin{tabular}{@{}c|c|c|c|c|c|c|c|c|c|c@{}}
\toprule
\multirow{2}{*}{Attack} &
  \textbf{Backdoored} &
  \textbf{FT} &
  \textbf{FP} \cite{liu2018fine} &
  \textbf{NAD} \cite{li2021neural} &
  \textbf{AC} \cite{chen2019detecting} &
  \textbf{NC} \cite{wang2019neural} &
  \textbf{ANP} \cite{wu2021adversarial} &
  \textbf{ABL} \cite{li2021anti} &
  \textbf{i-BAU} \cite{zeng2022adversarial} &
  \textbf{FT-SAM(Ours)} \\
 &
  ACC/ASR &
  ACC/ASR/DER &
  ACC/ASR/DER &
  ACC/ASR/DER &
  ACC/ASR/DER &
  ACC/ASR/DER &
  ACC/ASR/DER &
  ACC/ASR/DER &
  ACC/ASR/DER &
  ACC/ASR/DER \\ \hline \hline
BadNets-A2O\cite{gu2019badnets} &
  96.35/95.02 &
  \underline{97.60}/45.77/74.63 &
  \textbf{98.12}/\textbf{0.00}/\textbf{97.51} &
  97.54/79.94/57.54 &
  57.05/16.71/69.51 &
  93.47/0.02/96.06 &
  96.79/0.21/97.41 &
  94.53/\textbf{0.00}/96.60 &
  96.35/\textbf{0.00}/\textbf{97.51} &
  96.36/0.17/97.43 \\
BadNets-A2A\cite{gu2019badnets} &
  97.05/92.33 &
  \underline{98.04}/42.01/75.16 &
  \textbf{98.11}/0.51/\underline{95.91} &
  97.84/2.46/94.93 &
  96.14/80.93/55.25 &
  94.05/0.50/94.41 &
  96.73/50.39/70.81 &
  12.30/7.32/50.13 &
  95.30/\underline{0.43}/95.08 &
  96.97/\textbf{0.36}/\textbf{95.95} \\
Blended\cite{chen2017targeted} &
  97.97/99.67 &
  \underline{98.07}/94.09/52.79 &
  \textbf{98.31}/56.79/71.44 &
  97.76/95.90/51.78 &
  96.86/99.36/49.60 &
  88.04/\textbf{2.61}/\underline{93.57} &
  97.86/97.99/50.79 &
  43.29/4.66/70.17 &
  94.92/42.09/77.27 &
  96.55/\underline{3.13}/\textbf{97.56} \\
Input-aware\cite{nguyen2020input} &
  97.17/97.09 &
  97.58/47.14/74.97 &
  \underline{97.98}/1.36/\underline{97.86} &
  97.47/65.94/65.57 &
  38.43/51.69/43.33 &
  95.24/1.16/97.00 &
  96.20/\underline{1.12}/97.50 &
  9.97/59.83/25.03 &
  96.03/1.13/97.41 &
  \textbf{98.23}/\textbf{0.02}/\textbf{98.54} \\
LF\cite{zeng2021rethinking} &
  97.97/99.58 &
  98.00/83.83/57.88 &
  97.87/69.19/65.15 &
  \textbf{98.24}/79.76/59.91 &
  36.25/98.80/19.53 &
  92.22/\underline{0.18}/\underline{96.82} &
  \underline{98.03}/60.36/69.61 &
  26.29/0.68/63.61 &
  88.69/7.43/91.44 &
  96.52/\textbf{0.11}/\textbf{99.01} \\
SSBA\cite{li2021invisible} &
  98.31/99.77 &
  \underline{98.39}/98.88/50.45 &
  \textbf{98.47}/60.19/69.79 &
  98.37/96.95/51.41 &
  53.59/80.78/37.14 &
  90.75/1.51/\underline{95.35} &
  98.36/98.98/50.39 &
  50.89/\underline{0.50}/75.92 &
  87.27/\textbf{0.18}/94.27 &
  95.99/0.70/\textbf{98.37} \\
Trojan\cite{Trojannn} &
  98.33/100.00 &
  \textbf{98.38}/87.72/56.14 &
  98.00/42.08/78.80 &
  98.01/0.10/\textbf{99.79} &
  96.90/100.00/49.28 &
  92.29/0.02/96.97 &
  \underline{98.17}/86.92/56.46 &
  89.65/\textbf{0.00}/95.66 &
  93.66/\textbf{0.00}/97.66 &
  96.92/0.11/\underline{99.24} \\
Wanet\cite{nguyen2021wanet} &
  95.71/98.20 &
  \underline{98.69}/0.02/99.09 &
  \textbf{98.88}/0.28/98.96 &
  98.32/0.04/99.08 &
  61.67/2.14/81.01 &
  96.34/\underline{0.01}/\underline{99.09} &
  97.42/0.18/99.01 &
  40.36/86.25/28.30 &
  97.50/0.26/98.97 &
  98.61/\textbf{0.00}/\textbf{99.10} \\ \midrule
Avg &
  97.35/97.71 &
  \underline{98.09}/62.43/67.64 &
  \textbf{98.22}/28.80/84.43 &
  97.94/52.64/72.50 &
  67.11/66.30/50.58 &
  92.80/\underline{0.75}/\underline{96.16} &
  97.45/49.52/74.00 &
  45.91/19.91/63.18 &
  93.72/6.44/93.70 &
  97.02/\textbf{0.57}/\textbf{98.15} \\ \bottomrule
\end{tabular}%
\label{tablea1}
}
\end{table*}

%% file: tables_appendix/table2.tex
\begin{table*}[t]
\caption{Comparison with the SOTA defenses on \textbf{CIFAR-10} dataset with $5\%$ benign data on VGG19-BN (\%).}
\centering
\renewcommand\arraystretch{1.6} 
\resizebox{\textwidth}{!}{%
\begin{tabular}{@{}c|c|c|c|c|c|c|c|c|c|c@{}}
\toprule
\multirow{2}{*}{Attack} &
  \textbf{Backdoored} &
  \textbf{FT} &
  \textbf{FP} \cite{liu2018fine} &
  \textbf{NAD} \cite{li2021neural} &
  \textbf{AC} \cite{chen2019detecting} &
  \textbf{NC} \cite{wang2019neural} &
  \textbf{ANP} \cite{wu2021adversarial} &
  \textbf{ABL} \cite{li2021anti} &
  \textbf{i-BAU} \cite{zeng2022adversarial} &
  \textbf{FT-SAM(Ours)} \\
 &
  ACC/ASR &
  ACC/ASR/DER &
  ACC/ASR/DER &
  ACC/ASR/DER &
  ACC/ASR/DER &
  ACC/ASR/DER &
  ACC/ASR/DER &
  ACC/ASR/DER &
  ACC/ASR/DER &
  ACC/ASR/DER \\ \hline \hline
BadNets-A2O\cite{gu2019badnets} &
  90.42/94.43 &
  89.06/\underline{2.34}/\underline{95.36} &
  \underline{89.11}/12.39/90.37 &
  86.80/5.77/92.52 &
  84.79/93.01/47.90 &
  88.97/5.63/93.68 &
  \textbf{90.44}/87.64/53.39 &
  80.30/23.23/80.54 &
  87.69/3.13/94.29 &
  89.02/\textbf{1.52}/\textbf{95.76} \\
BadNets-A2A\cite{gu2019badnets} &
  91.16/84.39 &
  89.65/\textbf{1.09}/\textbf{90.90} &
  89.70/1.91/90.51 &
  88.15/1.60/89.89 &
  85.85/88.82/47.35 &
  \underline{91.16}/84.39/50.00 &
  \textbf{91.29}/81.87/51.26 &
  20.05/14.90/49.19 &
  86.86/2.19/88.95 &
  89.58/\underline{1.22}/\underline{90.80} \\
Blended\cite{chen2017targeted} &
  91.60/96.68 &
  \underline{89.66}/56.21/69.26 &
  89.54/72.33/61.14 &
  88.06/69.22/61.96 &
  86.72/99.98/47.56 &
  89.59/57.57/68.55 &
  \textbf{91.49}/91.04/52.76 &
  10.00/\textbf{0.00}/57.54 &
  87.17/9.22/\underline{91.51} &
  88.39/\underline{1.43}/\textbf{96.02} \\
Input-aware\cite{nguyen2020input} &
  88.66/94.58 &
  \textbf{91.34}/19.54/87.52 &
  \textbf{91.34}/5.42/94.58 &
  91.00/14.11/90.23 &
  48.01/22.54/65.69 &
  91.30/4.39/95.09 &
  89.67/20.43/87.07 &
  30.10/99.66/20.72 &
  88.30/\underline{3.70}/\underline{95.26} &
  90.59/\textbf{3.41}/\textbf{95.58} \\
CLA\cite{shafahi2018poison} &
  83.37/99.83 &
  88.56/8.42/95.71 &
  \textbf{88.80}/15.34/92.24 &
  87.39/\underline{7.83}/\underline{96.00} &
  78.91/97.16/49.11 &
  83.37/99.83/50.00 &
  83.24/57.31/71.20 &
  10.00/100.00/13.32 &
  85.68/11.23/94.30 &
  \textbf{88.80}/\textbf{7.50}/\textbf{96.17} \\
LF\cite{zeng2021rethinking} &
  83.28/13.83 &
  \underline{88.81}/1.31/56.26 &
  88.18/1.29/\underline{56.27} &
  85.08/3.07/55.38 &
  80.20/11.26/49.75 &
  88.33/\underline{1.22}/\textbf{56.31} &
  \textbf{89.20}/1.34/56.24 &
  55.30/\textbf{0.14}/42.85 &
  83.06/6.66/53.48 &
  88.45/1.79/56.02 \\
SIG\cite{barni2019new} &
  83.48/98.87 &
  88.11/2.90/97.98 &
  \textbf{88.66}/8.28/95.29 &
  86.14/6.30/96.28 &
  78.84/99.52/47.68 &
  83.48/98.87/50.00 &
  82.94/\textbf{0.00}/\textbf{99.16} &
  10.00/\textbf{0.00}/62.69 &
  84.50/4.47/97.20 &
  \underline{88.59}/2.00/\textbf{98.43} \\
SSBA\cite{li2021invisible} &
  90.85/95.11 &
  89.07/62.26/65.54 &
  89.26/65.33/64.09 &
  88.11/52.22/70.07 &
  85.81/90.63/49.72 &
  \underline{90.85}/95.11/50.00 &
  \textbf{91.11}/76.00/59.56 &
  10.00/\textbf{0.00}/57.13 &
  85.61/12.37/\underline{88.75} &
  89.25/\underline{3.30}/\textbf{95.11} \\
Trojan\cite{Trojannn} &
  91.57/100.00 &
  \underline{90.30}/6.63/96.05 &
  90.04/29.71/84.38 &
  87.01/5.17/95.14 &
  86.02/\underline{1.64}/\textbf{96.41} &
  \textbf{91.57}/100.00/50.00 &
  89.27/\textbf{0.00}/\textbf{98.85} &
  10.00/100.00/9.22 &
  86.40/2.69/96.07 &
  88.14/5.10/95.74 \\
Wanet\cite{nguyen2021wanet} &
  84.58/96.49 &
  \textbf{91.45}/2.79/96.85 &
  91.10/3.36/96.57 &
  90.68/10.23/93.13 &
  85.51/83.73/56.38 &
  84.58/96.49/50.00 &
  89.82/\textbf{0.96}/\textbf{97.77} &
  10.00/100.00/12.71 &
  89.61/2.40/97.05 &
  \underline{91.36}/\underline{1.00}/\textbf{97.75} \\ \midrule
Avg &
  87.90/87.42 &
  \textbf{89.60}/16.35/85.14 &
  \underline{89.57}/21.54/82.54 &
  87.84/17.55/84.06 &
  80.07/68.83/55.75 &
  88.32/64.35/61.36 &
  88.85/41.66/72.73 &
  24.58/43.79/40.59 &
  86.49/\underline{5.81}/\underline{89.69} &
  89.22/\textbf{2.83}/\textbf{91.74} \\ \bottomrule
\end{tabular}%
\label{tablea2}
}
\end{table*}

%% file: tables_appendix/table3.tex
\begin{table*}[]

\caption{Comparison with the SOTA defenses on \textbf{Tiny ImageNet} dataset with 5\% benign data on VGG19-BN (\%).}
\centering
\renewcommand\arraystretch{1.6} 
\resizebox{\textwidth}{!}{%
\begin{tabular}{@{}c|c|c|c|c|c|c|c|c|c|c@{}}
\toprule
\multirow{2}{*}{Attack} &
  \textbf{Backdoored} &
  \textbf{FT} &
  \textbf{FP} \cite{liu2018fine} &
  \textbf{NAD} \cite{li2021neural} &
  \textbf{AC} \cite{chen2019detecting} &
  \textbf{NC} \cite{wang2019neural} &
  \textbf{ANP} \cite{wu2021adversarial} &
  \textbf{ABL} \cite{li2021anti} &
  \textbf{i-BAU} \cite{zeng2022adversarial} &
  \textbf{FT-SAM(Ours)} \\
 &
  ACC/ASR &
  ACC/ASR/DER &
  ACC/ASR/DER &
  ACC/ASR/DER &
  ACC/ASR/DER &
  ACC/ASR/DER &
  ACC/ASR/DER &
  ACC/ASR/DER &
  ACC/ASR/DER &
  ACC/ASR/DER \\ \hline \hline
BadNets-A2O\cite{gu2019badnets} &
  43.56/99.96 &
  \underline{49.84}/99.45/50.26 &
  49.49/96.74/51.61 &
  49.35/0.27/\underline{99.84} &
  43.04/99.99/49.74 &
  43.57/99.96/50.00 &
  43.42/4.46/97.68 &
  41.10/\textbf{0.00}/98.75 &
  45.02/98.97/50.49 &
  \textbf{50.08}/\underline{0.14}/\textbf{99.91} \\
BadNets-A2A\cite{gu2019badnets} &
  54.44/50.74 &
  53.97/49.22/50.53 &
  53.13/\textbf{1.33}/\textbf{74.05} &
  \underline{54.13}/36.48/56.98 &
  42.98/36.57/51.36 &
  51.14/30.65/58.40 &
  \textbf{54.40}/\underline{1.99}/\textbf{74.36} &
  37.10/31.19/51.11 &
  46.72/36.46/53.28 &
  52.91/3.24/72.99 \\
Blended\cite{chen2017targeted} &
  50.68/97.08 &
  50.04/80.81/57.81 &
  49.78/64.10/66.04 &
  \underline{50.24}/57.45/69.59 &
  41.26/96.10/45.78 &
  48.84/\textbf{0.12}/\textbf{97.56} &
  \textbf{50.44}/95.46/50.69 &
  40.84/12.14/87.55 &
  45.57/89.55/51.21 &
  49.05/\underline{6.01}/\underline{94.72} \\
Input-aware\cite{nguyen2020input} &
  53.20/99.84 &
  53.33/\underline{0.06}/\textbf{99.89} &
  53.16/1.42/99.19 &
  \textbf{53.50}/0.14/99.85 &
  41.39/98.49/44.77 &
  53.29/0.08/99.88 &
  \underline{53.41}/\textbf{0.01}/\textbf{99.91} &
  40.48/3.48/91.82 &
  47.97/6.31/94.15 &
  51.78/0.26/99.08 \\
LF\cite{zeng2021rethinking} &
  48.92/7.73 &
  50.23/0.03/\underline{53.85} &
  50.29/0.02/\textbf{53.85} &
  \underline{50.44}/0.08/53.82 &
  39.28/9.90/45.18 &
  46.42/\textbf{0.01}/52.61 &
  \textbf{50.68}/0.39/53.67 &
  34.89/8.79/42.99 &
  43.81/0.03/51.29 &
  48.78/\underline{0.02}/53.78 \\
SSBA\cite{li2021invisible} &
  51.39/97.92 &
  50.58/88.93/54.09 &
  50.27/32.89/81.95 &
  50.23/71.66/62.55 &
  42.40/97.50/45.72 &
  49.39/\textbf{0.05}/\underline{97.93} &
  \underline{51.41}/97.26/50.33 &
  40.68/\underline{0.26}/93.47 &
  47.43/90.02/51.97 &
  \textbf{51.49}/1.70/\textbf{98.11} \\
Trojan\cite{Trojannn} &
  51.50/99.98 &
  50.94/98.84/50.29 &
  50.25/16.17/91.28 &
  \underline{51.02}/99.96/49.77 &
  42.92/99.90/45.75 &
  48.85/0.11/\underline{98.61} &
  \textbf{51.57}/97.06/51.46 &
  36.97/\textbf{0.00}/92.72 &
  43.77/99.69/46.28 &
  49.59/\underline{0.04}/\textbf{99.01} \\
Wanet\cite{nguyen2021wanet} &
  54.11/99.98 &
  \textbf{54.21}/0.14/\textbf{99.92} &
  53.69/19.59/89.99 &
  53.67/\textbf{0.10}/\underline{99.72} &
  41.14/96.03/45.49 &
  51.86/\underline{0.11}/98.81 &
  \underline{54.18}/60.06/69.96 &
  41.67/1.16/93.19 &
  48.32/88.33/52.93 &
  51.73/0.50/98.55 \\ \midrule
Avg &
  50.98/81.65 &
  \textbf{51.64}/52.19/64.58 &
  51.26/29.03/75.99 &
  \underline{51.57}/33.27/74.02 &
  41.80/79.31/46.72 &
  49.17/16.39/\underline{81.72} &
  51.19/44.58/68.51 &
  39.22/\underline{7.13}/81.45 &
  46.08/63.67/56.45 &
  50.68/\textbf{1.49}/\textbf{89.52} \\ \bottomrule
\end{tabular}%
\label{tablea3}
}
\end{table*}

%% file: tables_appendix/table4.tex
\begin{table*}[]
\caption{Comparison with the SOTA defenses on \textbf{GTSRB} dataset with 5\% benign data on VGG19-BN (\%).}
\centering
\renewcommand\arraystretch{1.6} 
\resizebox{\textwidth}{!}{%
\begin{tabular}{@{}c|c|c|c|c|c|c|c|c|c|c@{}}
\toprule
\multirow{2}{*}{Attack} &
  \textbf{Backdoored} &
  \textbf{FT} &
  \textbf{FP} \cite{liu2018fine} &
  \textbf{NAD} \cite{li2021neural} &
  \textbf{AC} \cite{chen2019detecting} &
  \textbf{NC} \cite{wang2019neural} &
  \textbf{ANP} \cite{wu2021adversarial} &
  \textbf{ABL} \cite{li2021anti} &
  \textbf{i-BAU} \cite{zeng2022adversarial} &
  \textbf{FT-SAM(Ours)} \\
 &
  ACC/ASR &
  ACC/ASR/DER &
  ACC/ASR/DER &
  ACC/ASR/DER &
  ACC/ASR/DER &
  ACC/ASR/DER &
  ACC/ASR/DER &
  ACC/ASR/DER &
  ACC/ASR/DER &
  ACC/ASR/DER \\ \hline \hline
BadNets-A2O\cite{gu2019badnets} &
  97.28/93.44 &
  97.42/30.37/81.54 &
  \textbf{97.63}/0.05/\textbf{96.69} &
  \underline{97.43}/89.78/51.83 &
  32.14/65.17/31.56 &
  94.78/\textbf{0.00}/95.47 &
  97.10/0.02/\underline{96.62} &
  3.56/\textbf{0.00}/49.86 &
  91.01/20.51/83.33 &
  95.98/0.03/96.05 \\
BadNets-A2A\cite{gu2019badnets} &
  97.59/93.29 &
  \textbf{98.40}/88.12/52.59 &
  \underline{98.34}/69.48/61.91 &
  97.83/88.16/52.57 &
  95.00/89.62/50.54 &
  95.34/1.09/94.98 &
  98.06/86.56/53.37 &
  10.55/8.30/48.97 &
  96.83/\underline{0.37}/\underline{96.08} &
  96.86/\textbf{0.20}/\textbf{96.18} \\
Blended\cite{chen2017targeted} &
  97.06/99.12 &
  \underline{97.43}/97.21/50.96 &
  \textbf{97.66}/97.30/50.91 &
  97.28/96.83/51.15 &
  95.31/98.38/49.50 &
  94.95/56.62/\underline{70.20} &
  97.13/98.56/50.28 &
  3.56/\textbf{0.00}/52.81 &
  96.38/63.49/67.48 &
  97.18/\underline{1.73}/\textbf{98.70} \\
Input-aware\cite{nguyen2020input} &
  96.32/85.03 &
  91.34/19.54/80.25 &
  \textbf{97.65}/0.49/92.27 &
  97.12/1.63/91.70 &
  31.88/17.28/51.65 &
  96.53/0.24/92.39 &
  96.94/\textbf{0.00}/\textbf{92.51} &
  1.74/81.95/4.25 &
  94.98/38.73/72.48 &
  \underline{97.25}/\underline{0.04}/\textbf{92.49} \\
LF\cite{zeng2021rethinking} &
  97.25/0.42 &
  97.07/0.03/50.10 &
  \textbf{97.59}/0.02/\textbf{50.20} &
  \underline{97.43}/0.02/\textbf{50.20} &
  28.04/3.88/15.40 &
  95.04/0.04/49.09 &
  97.35/0.45/50.00 &
  5.53/43.11/4.14 &
  89.75/\underline{0.01}/46.46 &
  95.19/\textbf{0.00}/49.18 \\
SSBA\cite{li2021invisible} &
  97.85/99.43 &
  \textbf{98.00}/98.97/50.23 &
  \underline{97.93}/98.77/50.33 &
  97.75/98.57/50.38 &
  31.61/71.26/30.97 &
  94.90/67.48/\underline{64.50} &
  97.85/99.34/50.04 &
  21.54/\textbf{0.00}/61.56 &
  86.98/99.92/44.57 &
  96.00/\underline{1.81}/\textbf{97.88} \\
Trojan\cite{Trojannn} &
  97.97/100.00 &
  97.68/8.27/95.72 &
  \textbf{98.00}/99.99/50.00 &
  97.76/6.34/96.73 &
  97.02/100.00/49.52 &
  95.61/0.02/98.82 &
  \underline{97.85}/97.28/51.30 &
  5.23/\textbf{0.00}/53.63 &
  96.01/\textbf{0.00}/\underline{99.02} &
  96.99/0.02/\textbf{99.51} \\
Wanet\cite{nguyen2021wanet} &
  94.76/98.32 &
  98.36/25.14/86.59 &
  \underline{98.66}/1.31/98.51 &
  98.37/0.20/99.06 &
  32.05/4.03/65.79 &
  96.41/7.30/95.51 &
  98.21/\underline{0.10}/\underline{99.11} &
  12.12/58.00/28.84 &
  87.17/10.30/90.21 &
  \textbf{98.76}/\textbf{0.04}/\textbf{99.14} \\ \midrule
Avg &
  97.01/83.63 &
  96.96/45.95/68.50 &
  \textbf{97.93}/45.93/68.85 &
  \underline{97.62}/47.69/67.95 &
  55.38/56.20/43.12 &
  95.45/\underline{16.60}/\underline{82.62} &
  97.56/47.79/67.90 &
  7.98/23.92/38.01 &
  92.39/29.17/74.95 &
  96.78/\textbf{0.48}/\textbf{91.14} \\ \bottomrule
\end{tabular}%
\label{tablea4} 
}
\end{table*}

%% file: tables_appendix/table5.tex
\begin{table*}[]
\caption{Comparison with the SOTA defenses with a \textbf{5\% poisoning ratio} on CIFAR-10 dataset with 5\% benign data on PreAct-ResNet18 (\%).}
\centering
\renewcommand\arraystretch{1.6} 
\resizebox{\textwidth}{!}{%
\begin{tabular}{@{}c|c|c|c|c|c|c|c|c|c|c@{}}
\toprule
\multirow{2}{*}{Attack} &
  \textbf{Backdoored} &
  \textbf{FT} &
  \textbf{FP} \cite{liu2018fine} &
  \textbf{NAD} \cite{li2021neural} &
  \textbf{AC} \cite{chen2019detecting} &
  \textbf{NC} \cite{wang2019neural} &
  \textbf{ANP} \cite{wu2021adversarial} &
  \textbf{ABL} \cite{li2021anti} &
  \textbf{i-BAU} \cite{zeng2022adversarial} &
  \textbf{FT-SAM(Ours)} \\
 &
  ACC/ASR &
  ACC/ASR/DER &
  ACC/ASR/DER &
  ACC/ASR/DER &
  ACC/ASR/DER &
  ACC/ASR/DER &
  ACC/ASR/DER &
  ACC/ASR/DER &
  ACC/ASR/DER &
  ACC/ASR/DER \\ \hline \hline
BadNets-A2O\cite{gu2019badnets} &
  92.35/89.52 &
  90.83/2.50/92.75 &
  92.10/1.47/\textbf{93.90} &
  89.92/1.98/92.56 &
  88.67/88.33/48.75 &
  90.88/1.62/93.22 &
  \underline{92.23}/2.80/93.30 &
  81.58/\textbf{0.00}/89.38 &
  89.61/\underline{1.00}/92.89 &
  \textbf{92.27}/2.12/\underline{93.66} \\
BadNets-A2A\cite{gu2019badnets} &
  92.54/65.85 &
  91.78/\underline{0.93}/82.08 &
  \underline{92.37}/1.02/\underline{82.33} &
  91.19/1.38/81.56 &
  87.71/54.64/53.19 &
  89.46/1.25/80.76 &
  91.93/1.45/81.90 &
  42.31/38.38/38.62 &
  90.69/1.67/81.17 &
  \textbf{92.54}/\textbf{0.91}/\textbf{82.47} \\
Blended\cite{chen2017targeted} &
  93.66/94.82 &
  \underline{93.18}/83.63/55.35 &
  93.10/9.64/\underline{92.31} &
  93.08/66.46/63.89 &
  89.27/87.52/51.46 &
  93.00/87.53/53.31 &
  \textbf{93.24}/82.26/56.07 &
  73.23/\textbf{0.19}/87.10 &
  86.73/\underline{1.30}/\textbf{93.30} &
  91.07/8.27/91.98 \\
Input-aware\cite{nguyen2020input} &
  91.51/93.05 &
  93.08/66.97/63.04 &
  93.17/26.71/83.17 &
  \underline{93.28}/92.26/50.40 &
  89.05/72.60/59.00 &
  93.23/82.31/55.37 &
  91.06/\underline{13.31}/\underline{89.65} &
  85.54/83.97/51.56 &
  91.28/22.10/85.36 &
  \textbf{93.69}/\textbf{6.23}/\textbf{93.41} \\
CLA\cite{shafahi2018poison} &
  93.47/99.33 &
  92.67/96.29/51.12 &
  92.38/39.00/79.62 &
  92.38/90.19/54.03 &
  89.87/96.14/49.79 &
  \textbf{93.47}/99.33/50.00 &
  92.76/\underline{23.16}/\underline{87.73} &
  73.52/99.67/40.03 &
  88.26/40.60/76.76 &
  \underline{92.86}/\textbf{5.70}/\textbf{96.51} \\
LF\cite{zeng2021rethinking} &
  93.51/97.29 &
  \textbf{93.19}/96.23/50.37 &
  92.11/69.07/63.41 &
  92.93/94.96/50.88 &
  89.12/95.33/48.78 &
  \underline{93.04}/54.28/71.27 &
  93.01/73.98/61.41 &
  61.19/94.11/35.43 &
  89.85/\underline{28.73}/\underline{82.45} &
  92.74/\textbf{3.81}/\textbf{96.35} \\
SIG\cite{barni2019new} &
  93.29/95.06 &
  92.73/92.41/51.04 &
  \underline{92.87}/43.99/75.32 &
  92.21/82.62/55.68 &
  89.66/94.44/48.49 &
  \textbf{93.29}/95.06/50.00 &
  92.78/97.47/49.75 &
  57.72/\textbf{0.00}/79.74 &
  88.04/7.30/\underline{91.25} &
  92.62/\underline{0.61}/\textbf{96.89} \\
SSBA\cite{li2021invisible} &
  93.08/94.09 &
  92.62/83.63/55.00 &
  92.23/13.70/89.77 &
  92.35/86.03/53.66 &
  89.12/86.92/51.60 &
  \textbf{93.08}/94.09/50.00 &
  \underline{93.07}/79.38/57.35 &
  78.75/\textbf{0.94}/89.41 &
  90.62/\underline{2.62}/\underline{94.50} &
  92.35/3.84/\textbf{94.76} \\
Trojan\cite{Trojannn} &
  93.61/99.99 &
  92.82/99.87/49.67 &
  92.77/88.68/55.24 &
  93.08/31.86/83.80 &
  89.61/99.97/48.01 &
  93.03/99.79/49.81 &
  \textbf{93.26}/99.99/49.83 &
  70.19/\textbf{0.00}/88.28 &
  89.19/\underline{4.89}/\underline{95.34} &
  \underline{93.12}/6.84/\textbf{96.33} \\
Wanet\cite{nguyen2021wanet} &
  93.38/97.27 &
  \textbf{93.45}/19.96/88.65 &
  93.01/1.50/97.70 &
  93.31/8.56/94.32 &
  88.13/58.24/66.89 &
  \underline{93.38}/97.27/50.00 &
  92.93/\textbf{0.31}/\textbf{98.26} &
  60.52/99.04/33.57 &
  89.16/1.58/95.74 &
  93.27/\underline{0.80}/\textbf{98.18} \\ \midrule
Avg &
  93.04/92.63 &
  \underline{92.64}/64.24/63.91 &
  92.61/29.48/81.28 &
  92.37/55.63/68.08 &
  89.02/83.41/52.60 &
  92.59/71.25/60.37 &
  92.63/47.41/72.52 &
  68.46/41.63/63.31 &
  89.34/\underline{11.18}/\underline{88.88} &
  \textbf{92.65}/\textbf{3.91}/\textbf{94.05} \\ \bottomrule
\end{tabular}%
\label{tablea5} 
}
\end{table*}

%% file: tables_appendix/table6.tex
\begin{table*}[]
\caption{Comparison with the SOTA defenses with a \textbf{1\% poisoning ratio} on CIFAR-10 dataset with 5\% benign data on PreAct-ResNet18 (\%).}
\centering
\renewcommand\arraystretch{1.6} 
\resizebox{\textwidth}{!}{%
\begin{tabular}{@{}c|c|c|c|c|c|c|c|c|c|c@{}}
\toprule
\multirow{2}{*}{Attack} &
  \textbf{Backdoored} &
  \textbf{FT} &
  \textbf{FP} \cite{liu2018fine} &
  \textbf{NAD} \cite{li2021neural} &
  \textbf{AC} \cite{chen2019detecting} &
  \textbf{NC} \cite{wang2019neural} &
  \textbf{ANP} \cite{wu2021adversarial} &
  \textbf{ABL} \cite{li2021anti} &
  \textbf{i-BAU} \cite{zeng2022adversarial} &
  \textbf{FT-SAM(Ours)} \\
 &
  ACC/ASR &
  ACC/ASR/DER &
  ACC/ASR/DER &
  ACC/ASR/DER &
  ACC/ASR/DER &
  ACC/ASR/DER &
  ACC/ASR/DER &
  ACC/ASR/DER &
  ACC/ASR/DER &
  ACC/ASR/DER \\ \hline \hline
BadNets-A2O\cite{gu2019badnets} &
  93.12/74.20 &
  90.83/2.50/84.71 &
  \underline{92.88}/2.44/\underline{85.76} &
  92.38/10.87/81.30 &
  89.25/11.84/79.24 &
  92.77/30.67/71.59 &
  \textbf{93.09}/5.84/84.16 &
  82.56/\textbf{0.83}/81.40 &
  90.05/2.80/84.17 &
  92.74/\underline{1.31}/\textbf{86.25} \\
BadNets-A2A\cite{gu2019badnets} &
  93.42/28.62 &
  91.78/\textbf{0.93}/63.03 &
  92.32/\underline{1.07}/\underline{63.23} &
  92.28/1.87/62.81 &
  88.72/1.86/61.03 &
  \textbf{93.42}/28.60/50.01 &
  \underline{93.12}/4.78/61.77 &
  52.76/21.93/33.02 &
  88.84/2.11/60.97 &
  92.74/1.11/\textbf{63.42} \\
Blended\cite{chen2017targeted} &
  93.69/73.88 &
  93.18/83.63/49.75 &
  92.99/\underline{5.97}/\underline{83.61} &
  93.24/47.94/62.74 &
  89.62/41.51/64.15 &
  \textbf{93.69}/73.88/50.00 &
  \underline{93.25}/49.33/62.05 &
  74.45/24.90/64.87 &
  86.56/7.26/79.75 &
  91.85/\textbf{4.62}/\textbf{83.71} \\
Input-aware\cite{nguyen2020input} &
  91.15/68.53 &
  93.08/66.97/50.78 &
  93.07/20.81/73.86 &
  93.13/84.92/50.00 &
  89.99/56.44/55.46 &
  \underline{93.19}/57.83/55.35 &
  91.58/63.79/52.37 &
  58.76/21.92/57.11 &
  90.72/\underline{10.62}/\underline{78.74} &
  \textbf{93.31}/\textbf{1.47}/\textbf{83.53} \\
CLA\cite{shafahi2018poison} &
  93.71/94.41 &
  92.67/96.29/49.48 &
  93.03/30.77/81.48 &
  93.31/87.73/53.14 &
  89.58/11.34/\underline{89.47} &
  \underline{93.38}/91.00/51.54 &
  \textbf{93.50}/93.83/50.18 &
  65.80/14.00/76.25 &
  88.67/\underline{11.01}/89.18 &
  92.96/\textbf{4.97}/\textbf{94.35} \\
LF\cite{zeng2021rethinking} &
  93.29/85.94 &
  \textbf{93.19}/96.23/49.95 &
  92.31/61.76/61.60 &
  92.91/77.48/54.04 &
  89.38/78.67/51.68 &
  91.09/\textbf{3.64}/\underline{90.05} &
  \underline{93.08}/45.53/70.10 &
  56.17/63.32/42.75 &
  85.67/72.01/53.16 &
  92.10/\underline{4.50}/\textbf{90.13} \\
SIG\cite{barni2019new} &
  93.68/78.68 &
  92.73/92.41/49.53 &
  92.02/67.74/54.64 &
  93.13/78.11/50.01 &
  90.12/79.77/48.22 &
  \textbf{93.68}/78.68/50.00 &
  \underline{93.47}/78.38/50.05 &
  65.12/\textbf{0.00}/\underline{75.06} &
  90.11/31.69/71.71 &
  91.43/\underline{3.29}/\textbf{86.57} \\
SSBA\cite{li2021invisible} &
  93.51/70.69 &
  92.62/83.63/49.56 &
  \underline{93.17}/7.20/81.57 &
  93.15/54.54/57.89 &
  89.29/31.38/67.55 &
  93.16/54.89/57.73 &
  \textbf{93.28}/24.48/72.99 &
  59.42/65.03/35.78 &
  90.42/\textbf{1.10}/\underline{83.25} &
  92.96/\underline{1.81}/\textbf{84.16} \\
Trojan\cite{Trojannn} &
  93.80/99.89 &
  92.82/99.87/49.52 &
  92.91/98.32/50.34 &
  \underline{93.45}/99.87/49.84 &
  89.95/99.73/48.15 &
  93.42/99.91/49.81 &
  \textbf{93.51}/99.86/49.87 &
  61.68/\underline{43.73}/62.02 &
  87.56/59.07/\underline{67.29} &
  93.14/\textbf{8.23}/\textbf{95.50} \\
Wanet\cite{nguyen2021wanet} &
  93.03/81.05 &
  \textbf{93.45}/19.96/80.54 &
  \underline{93.33}/\textbf{0.49}/\textbf{90.28} &
  93.27/2.59/89.23 &
  89.18/4.67/86.27 &
  93.21/3.51/88.77 &
  92.75/1.24/89.77 &
  29.86/81.91/18.42 &
  90.64/1.19/88.73 &
  93.21/\underline{0.76}/\underline{90.15} \\ \midrule
Avg &
  93.24/75.59 &
  92.64/64.24/57.68 &
  92.80/29.66/72.64 &
  93.03/54.59/61.10 &
  89.51/41.72/65.12 &
  \textbf{93.10}/52.26/61.48 &
  \underline{93.06}/46.71/64.33 &
  60.66/33.76/54.67 &
  88.92/\underline{19.89}/\underline{75.69} &
  92.64/\textbf{3.21}/\textbf{85.78} \\ \bottomrule
\end{tabular}%
\label{tablea6}
}
\end{table*}

%% file: tables_appendix/table9.tex
\begin{table*}[]
\caption{Comparison with the state-of-the-art defenses on CIFAR-10 dataset with 5\% benign data on PreAct-ResNet18 (\%). The better result between the two is indicated in \textbf{boldface}.}
\centering
\renewcommand\arraystretch{1.6} 
\resizebox{\textwidth}{!}{%
\begin{tabular}{@{}c|c|c|c|c|c|c|c|c|c|c|c@{}}
\toprule
\multirow{2}{*}{Model} &
  BadNets-A2O\cite{gu2019badnets} &
  BadNets-A2A\cite{gu2019badnets} &
  Blended\cite{chen2017targeted} &
  Input-aware\cite{nguyen2020input} &
  CLA\cite{shafahi2018poison} &
  LF\cite{zeng2021rethinking} &
  SIG\cite{barni2019new} &
  SSBA\cite{li2021invisible} &
  Trojan\cite{Trojannn} &
  Wanet\cite{nguyen2021wanet} &
  Avg \\
 &
  ACC/ASR/DER &
  ACC/ASR/DER &
  ACC/ASR/DER &
  ACC/ASR/DER &
  ACC/ASR/DER &
  ACC/ASR/DER &
  ACC/ASR/DER &
  ACC/ASR/DER &
  ACC/ASR/DER &
  ACC/ASR/DER &
  ACC/ASR/DER \\ \hline \hline
Backdoored &
  91.82/93.79/- &
  91.89/74.42/- &
  93.44/97.71/- &
  94.03/98.35/- &
  84.55/99.93/- &
  93.01/99.06/- &
  84.49/97.87/- &
  92.88/97.07/- &
  93.47/99.99/- &
  92.80/98.90/- &
  91.24/95.71/- \\
\textbf{w/o Adaptive} &
  90.85/\textbf{1.53}/95.64 &
  90.95/1.39/86.05 &
  91.30/\textbf{2.44}/\textbf{96.56} &
  92.94/1.39/97.93 &
  89.95/6.19/96.87 &
  90.77/6.73/95.04 &
  89.84/\textbf{0.49}/\textbf{98.69} &
  90.74/5.78/94.57 &
  90.80/14.02/91.65 &
  91.94/1.89/98.07 &
  91.01/4.19/95.65 \\
\textbf{w/ Adaptive} &
  \textbf{92.21}/1.63/\textbf{96.08} &
  \textbf{91.87}/\textbf{1.03}/\textbf{86.69} &
  \textbf{92.44}/4.91/95.90 &
  \textbf{93.76}/\textbf{1.07}/\textbf{98.51} &
  \textbf{90.72}/\textbf{3.52}/\textbf{98.21} &
  \textbf{91.07}/\textbf{3.81}/\textbf{96.65} &
  \textbf{91.16}/0.80/98.53 &
  \textbf{92.12}/\textbf{2.80}/\textbf{96.75} &
  \textbf{92.75}/\textbf{4.12}/\textbf{97.57} &
  \textbf{92.87}/\textbf{0.96}/\textbf{98.97} &
  \textbf{92.10}/\textbf{2.47}/\textbf{96.62} \\ \bottomrule
\end{tabular}%
\label{tablea9}
}
\end{table*}

%% file: tables_appendix/table10.tex
\begin{table*}[!t]
\caption{Defense results of $l_2$ weight norm regularization on CIFAR-10 dataset with 5\% benign data on PreAct-ResNet18 (\%).}
\centering
\renewcommand\arraystretch{1.3} 
\resizebox{0.8\textwidth}{!}{%
\begin{tabular}{@{}c|c|c|c|c|c|c@{}}
\toprule
Attack &
  BadNets-A2O\cite{gu2019badnets} &
  Blended\cite{chen2017targeted} &
  Input-aware\cite{nguyen2020input} &
  LF\cite{zeng2021rethinking} &
  SSBA\cite{li2021invisible} &
  Trojan\cite{Trojannn} \\
$\gamma$   & ACC/ASR/DER      & ACC/ASR/DER       & ACC/ASR/DER       & ACC/ASR/DER       & ACC/ASR/DER       & ACC/ASR/DER       \\ \hline\hline
(Attack)    & 91.82/93.79/-    & 93.44/97.71/-     & 94.03/98.35/-     & 93.01/99.06/-     & 84.49/97.87/-     & 92.88/97.07/-     \\

0.001 & 90.69/1.27/\textbf{95.70} & 92.77/74.13/61.45 & 93.97/10.55/\textbf{89.84} & 92.40/85.36/56.55 & 92.37/67.56/65.16 & 92.70/26.38/85.25 \\
0.005 & 89.24/1.33/94.94 & 91.80/7.73/94.17  & 93.67/10.76/89.73 & 91.75/33.13/82.33 & 91.69/9.24/94.31  & 92.23/11.03/\textbf{92.69} \\
0.01  & 88.99/0.80/95.08 & 89.13/1.24/\textbf{96.08}  & 92.07/11.06/89.58 & 90.54/13.59/\textbf{91.50} & 89.04/2.58/\textbf{97.64}  & 89.62/9.98/91.91  \\
0.05  & 36.55/0.11/69.20 & 28.44/1.59/65.56  & 41.08/6.87/67.47  & 35.85/17.08/62.41 & 49.13/2.18/80.16  & 44.50/9.27/69.71  \\
0.1   & 18.47/2.59/58.93 & 12.79/6.41/55.33  & 18.90/3.10/58.27  & 17.93/2.27/60.85  & 10.10/0.51/61.48  & 13.66/2.87/57.49  \\ \bottomrule
\end{tabular}%
\label{tablea10}
}
\end{table*}